\documentclass[letterpaper]{article} %
\usepackage{aaai23}  %
\usepackage{times}  %
\usepackage{helvet}  %
\usepackage{courier}  %
\usepackage[hyphens]{url}  %
\usepackage{graphicx} %
\usepackage{natbib}  %
\usepackage{caption} %
\usepackage{algorithm}
\usepackage{algorithmic}

\usepackage{newfloat}
\usepackage{listings}
\DeclareCaptionStyle{ruled}{labelfont=normalfont,labelsep=colon,strut=off} %
\floatstyle{ruled}
\newfloat{listing}{tb}{lst}{}
\floatname{listing}{Listing}
\usepackage{color} %

\title{Efficient Mirror Detection via Multi-level Heterogeneous Learning}
\author{
    Ruozhen He, \ \ 
    Jiaying Lin\thanks{Corresponding authors: Jiaying Lin and Rynson W.H. Lau}, \ \
    Rynson W.H. Lau$^*$
}
\affiliations{
    Department of Computer Science, City University of Hong Kong\\
    ruozhenhe2-c@my.cityu.edu.hk,
    jiayinlin5-c@my.cityu.edu.hk,
    rynson.lau@cityu.edu.hk

}

\usepackage{bibentry}

\usepackage{mathrsfs}
\usepackage{amsmath}

\usepackage[dvipsnames]{xcolor}
\usepackage{amssymb}
\definecolor{mygreen}{RGB}{0,0,0}
\definecolor{myyellow}{RGB}{0, 0, 0}
\newcommand{\jy}[1]{\textcolor{mygreen}{#1}}

\newcommand{\rz}[1]{\textcolor{black}{#1}}
\newcommand{\ryn}[1]{{#1}}

\begin{document}

\maketitle

\begin{abstract}
We present HetNet (Multi-level \textbf{Het}erogeneous \textbf{Net}work), a highly efficient mirror detection network. Current mirror detection methods focus more on performance than efficiency, limiting the real-time applications (such as drones). Their lack of efficiency is aroused by the common design of adopting homogeneous modules at different levels, which ignores the difference between different levels of features.
In contrast, HetNet detects potential mirror regions initially through low-level understandings (\textit{e.g.}, intensity contrasts) and then combines with high-level understandings (contextual discontinuity for instance) to finalize the predictions. 
To perform accurate yet efficient mirror detection, HetNet follows an effective architecture that obtains specific information at different stages to detect mirrors. We further propose a multi-orientation intensity-based contrasted module (MIC) and a reflection semantic logical module (RSL), equipped on HetNet,  to predict potential mirror regions by low-level understandings and analyze semantic logic in scenarios by high-level understandings, respectively. 
Compared to the state-of-the-art method, HetNet runs 664$\%$ faster and draws an average performance gain of 8.9$\%$ on MAE, 3.1$\%$ on IoU, and 2.0$\%$ on F-measure on two mirror detection benchmarks.  
The code is available at \textit{\url{https://github.com/Catherine-R-He/HetNet}}.
\end{abstract}

\section{Introduction}

Mirrors are common objects in our daily lives. The reflection of mirrors may cause depth prediction errors and confusion about reality and virtuality. Ignoring them in computer vision tasks may cause severe safety issues \ryn{in situation such as drone and robotic navigation}. In addition, owing to the limited computation resources, application scenarios \ryn{may heavily depend on the model efficiency while} efficient mirror detection is essential for real-time computer vision applications.

\begin{figure}[t]
    \centering
    \includegraphics[width=0.5\textwidth]{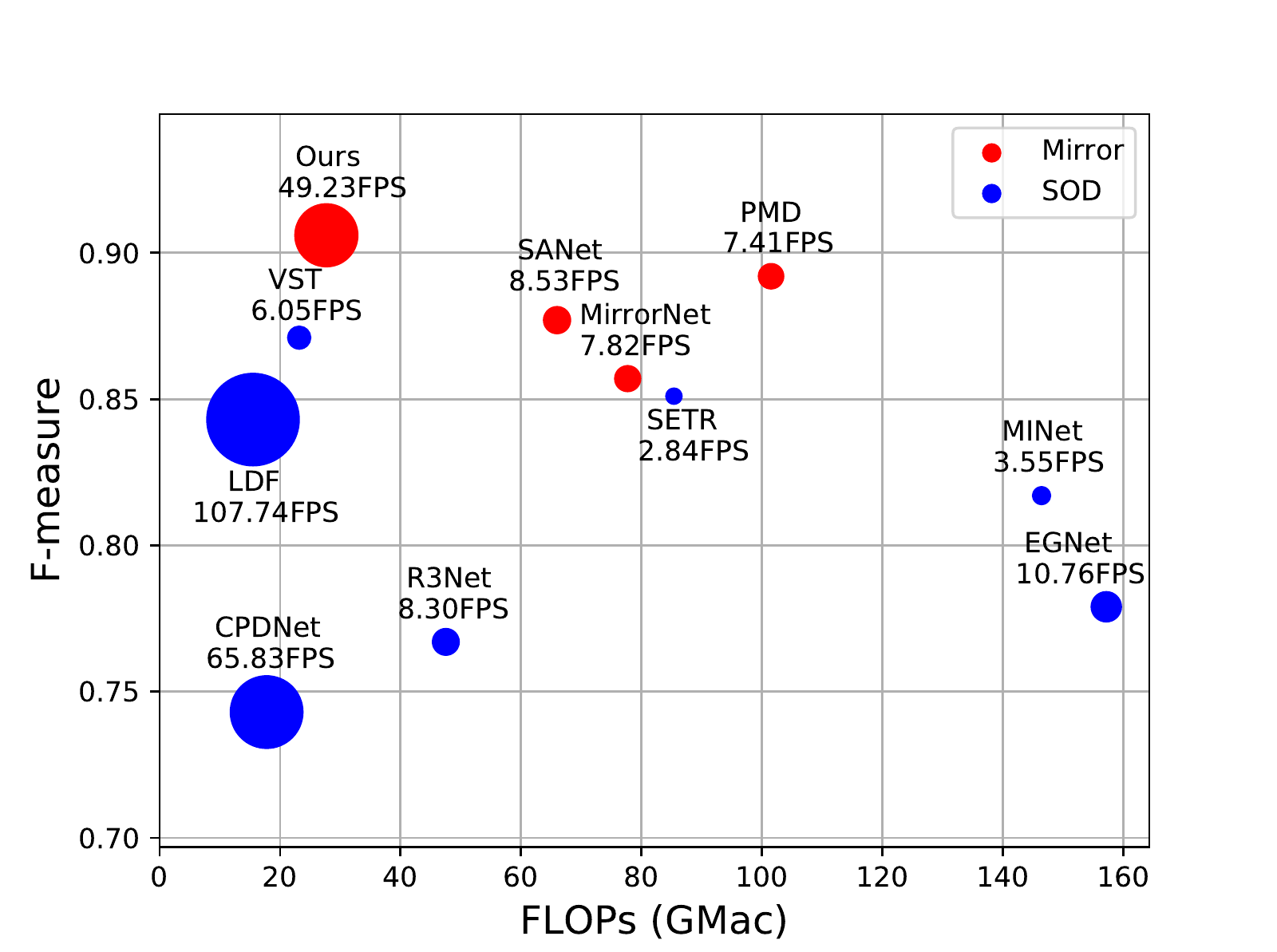}
    \caption{Comparison of our proposed method with Mirror Detection and SOD models on F-measure, FLOPs (GMac), and FPS using the MSD dataset. Our method achieves a new SOTA result with considerable efficiency.}
    \label{fig:flops}
\end{figure}

Recently, Yang \textit{et al.} \cite{yang2019my} propose MirrorNet based on contextual contrasted features. Lin \textit{et al.} \cite{lin2020progressive} propose PMD, \ryn{which considers} content similarity. Guan \textit{et al.} \cite{guan2022learning} propose SANet, \ryn{which focuses} on semantic associations. Though experimental results show their superior performances on mirror detection, these methods suffer from huge computation costs since they apply the same modules for both low-level features with large spatial resolutions and high-level features with small spatial resolutions at every stage. Besides, they heavily rely on post-processing algorithms, \textit{e.g.,} CRF~\cite{krahenbuhl2011efficient}, \ryn{which} heavily limit the usage of these existing methods to real-world scenarios with the demand for real-time processing, \ryn{as demonstrated in Figure~\ref{fig:flops}}. 
For more comparisons, we also \ryn{include} some methods from a related task, salient object detection (SOD). Although some of them (\textit{e.g.,} LDF~\cite{wei2020label}, CPDNet~\cite{wu2019cascaded}) contain few FLOPs with high FPS, they \ryn{are not able to achieve competitive performances}. Thus, it is challenging and significant to propose a method \ryn{that meets the trade-off between} accuracy and efficiency.

\begin{figure}[t]
    \centering
    \includegraphics[width = 0.48\textwidth]{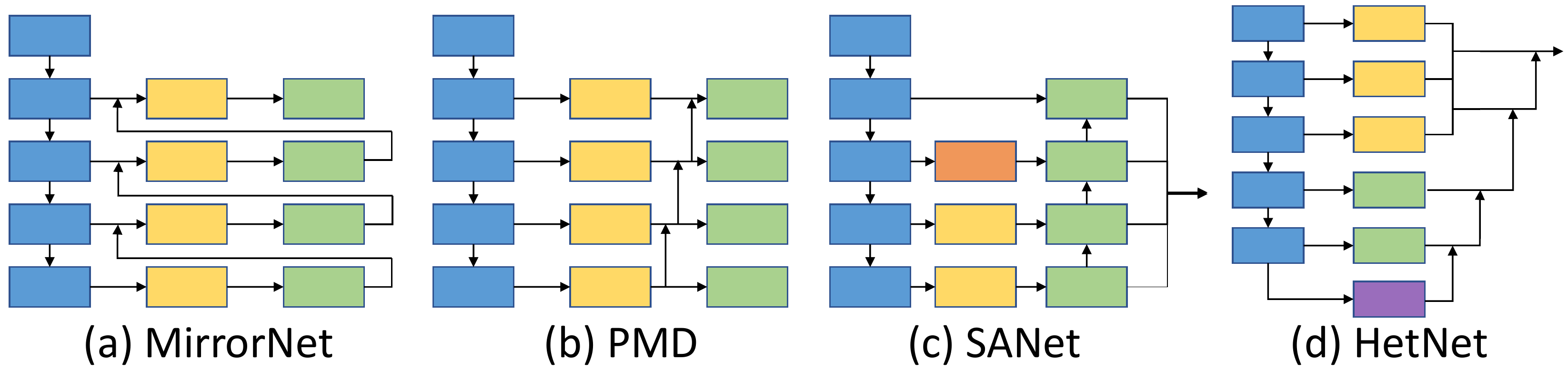}
    \caption{Illustrations of different network architectures used in the existing \ryn{mirror detection methods}. (a) to (d) is the network architecture of MirrorNet~\cite{yang2019my}, PMD~\cite{lin2020progressive}, SANet~\cite{guan2022learning}, and HetNet, respectively.
    We denote different modules in different colors. \ryn{Both} MirrorNet and PMD share the same high-level design, which uses the same type of modules to learn from the backbone features at different levels, while SANet aggregates all-level features simultaneously. Different from these designs, we split the backbone features as low- and high-level features, and adopt different modules to learn them. For example, in (d), the low-level features are fed into multi-orientation intensity-based contrasted modules (in yellow),
    and the high-level features are forwarded into reflection semantic logical modules (in green).
    Such design can fully exploit features at different levels effectively and efficiently.
    }
    \label{fig:arch}
\end{figure}

Figure~\ref{fig:arch} \ryn{compares} different network architectures for mirror detection. 
\jy{As shown in Figure~\ref{fig:arch}(a-c), existing mirror detection network architectures use the same module for both low-level \ryn{and} high-level features~\cite{yang2019my,lin2020progressive,guan2022learning}.}
However, although low-level features contain the representation of colors, shapes, and textures, learning \ryn{these features require a higher computational} cost due to their larger spatial resolutions compared with high-level features.
On the other hand, high-level features involve more semantic information, \ryn{and it is more} difficult for models to directly extract precise mirror boundaries as \ryn{high-level features contain only rough spatial information.}
On account of the representation gap between low-level and high-level features, it is inappropriate to adopt the same module
for both \ryn{types of features in model} design.

To overcome these limitations, we propose in this paper a highly efficient HetNet (Multi-level \textbf{Het}erogeneous \textbf{Net}work). We observe that both low-level and high-level understandings assist mirror detection. \ryn{We also} notice that cells on the retina convert the received light signals into electrical signals \cite{wald1935carotenoids}, which are then processed and transmitted to the visual cortex for further information processing, integration and abstraction \cite{hubel1962receptive}, thereby forming vision. HetNet mimics this process in mirror detection. Specifically, human eyes initially accept low-level information (\textit{e.g.}, intensity contrast, colors) \ryn{from the scene}, and after transmission and abstraction (\textit{e.g.}, objects' edges and shapes) in our brains, we confirm the reflection semantic high-level understandings (\textit{e.g.}, content similarity, contextual contrast) of objects to finally determine mirrors. Considering these observations, we propose our new model HetNet \ryn{to take advantage of the characteristics of low-level and high-level features individually. HetNet includes multi-orientation intensity-based contrasted (MIC) modules to learn low-level features at shallow stages to help localize mirrors, and reflection semantic logical (RSL) modules to help extract high-level understandings at deep stages and then aggregate them with low-level understandings to output the final mirror mask}. 
In addition, we fully use the backbone network for better learning of low-level features without \ryn{a huge computational cost}.
With the benefit of the disentangled learning on low-level and high-level features, our model exploits low-level and high-level features effectively and efficiently. Experimental results show that with the proper heterogeneous design for features of different levels, our model outperforms the state-of-the-art mirror detection method PMD~\cite{lin2020progressive} with 72.73$\%$ fewer model FLOPs and 664$\%$ faster.

Our main contributions are summarized as follows:

\begin{itemize}
    \item[$\bullet$]We propose the first highly efficient mirror detection model HetNet which learns specific understandings via heterogeneous modules at different levels. \jy{Unlike existing mirror detection methods that adopt the same modules in all stages, such\ryn{our heterogeneous network can reduce the computational cost via a proper design for different levels of} features.}
    \item[$\bullet$]We propose a novel model that consists of multi-orientation intensity-based contrasted (MIC) modules for initial localization by low-level features in multi-orientation, and reflection semantic logical (RSL) modules for semantic analysis through high-level features.
    \item[$\bullet$]Experiments demonstrate that HetNet outperforms all relevant baselines, achieving outstanding efficiency (664$\%$ faster) and accuracy (an average enhancement of 8.9$\%$ on MAE, 3.1$\%$ on IoU, and 2.0$\%$ on F-measure on two benchmarks) compared with the SOTA method PMD. 
\end{itemize}

\section{Related Works}

\subsection{Mirror Detection}
The initial work of automatic mirror detection was proposed by Yang \textit{et al.} \cite{yang2019my}. They localize mirrors through multi-scale contextual contrasting features. Thus, this method has limitations if the mirror and non-mirror contents are similar. To solve the problem, Lin\textit{ et al.} \cite{lin2020progressive} propose a method focused on the relationship between features of mirror and non-mirror areas. Recently, two concurrent works~\cite {tan2022mirror} and \cite{huang2023symmetry} adopt heavy structures (e.g., transformer) for mirror detection despite low efficiency.
However, these methods fail when an area is likely to be a mirror from a context aspect. They are also not efficient for real-time mirror detection. 

To overcome the limitation, our method preliminary localizes mirror regions based on the intensity contrast. Then it integrates contextual relationships inside and outside mirrors to refine the final predicted regions. Experiments results prove our approach has better performance on both MSD \cite{yang2019my} and PMD \cite{lin2020progressive} datasets.

\subsection{Salient Object Detection}
It detects the most salient objects in images. Early methods are mainly based on low-level features such as color and contrast \cite{achanta2009frequency} and spectral residual \cite{hou2007saliency}. Recently, most approaches have depended on deep learning. Qin \textit{et al.} \cite{qin2019basnet} propose a densely supervised encoder-decoder with a residual refine module to generate and refine saliency maps. Chen \textit{et al.} \cite{chen2020global} propose a global context-aware progressive aggregation network to aggregate multi-level features. Pang \textit{et al.} \cite{pang2020multi} design an aggregate interaction module to extract useful inter-layer features via interactive learning. Ma \textit{et al.} \cite{ma2021pyramidal} extract effective features and denoise through an adjacent fusion module. Liu \textit{et al.} \cite{liu2021visual} propose a transformer-based model from a sequence-to-sequence perspective.
Nevertheless, the mirror reflects a part of a scene, including both salient and non-salient objects. Hence, salient object detection may not detect mirrors precisely.

\subsection{Shadow Detection}

It identifies or removes shadow areas of images. Nguyen \textit{et al.} \cite{nguyen2017shadow} propose a conditional generative adversarial network supporting multi-sensitivity level shadow generation. Hu \textit{et al.} \cite{hu2018direction} propose a direction-aware spatial context module based on spatial RNN to learn spatial contexts. Zhu \textit{et al.} \cite{zhu2018bidirectional} refine context features recurrently through a recurrent attention residual module. Zheng \textit{et al.} \cite{zheng2019distraction} design a distraction-aware shadow module to solve indistinguishable area problems. Zhu \textit{et al. }\cite{zhu2021mitigating} propose a feature decomposition and reweighting to adjust the significance of intensity and other features. Han \textit{et al.} \cite{han2022bid} further incorporate shadow detection in a blind image decomposition setting. 
The main factor of shadow detection is the distinct intensity contrast between non-shadow and shadow areas. However, there are usually no strong intensity contrasts in the mirror detection scene but many weak ones. Therefore, it is hard to detect mirrors by shadow detection methods.

\section{Methodology}
\begin{figure*}[h]
    \centering
    \includegraphics[width=0.75\textwidth]{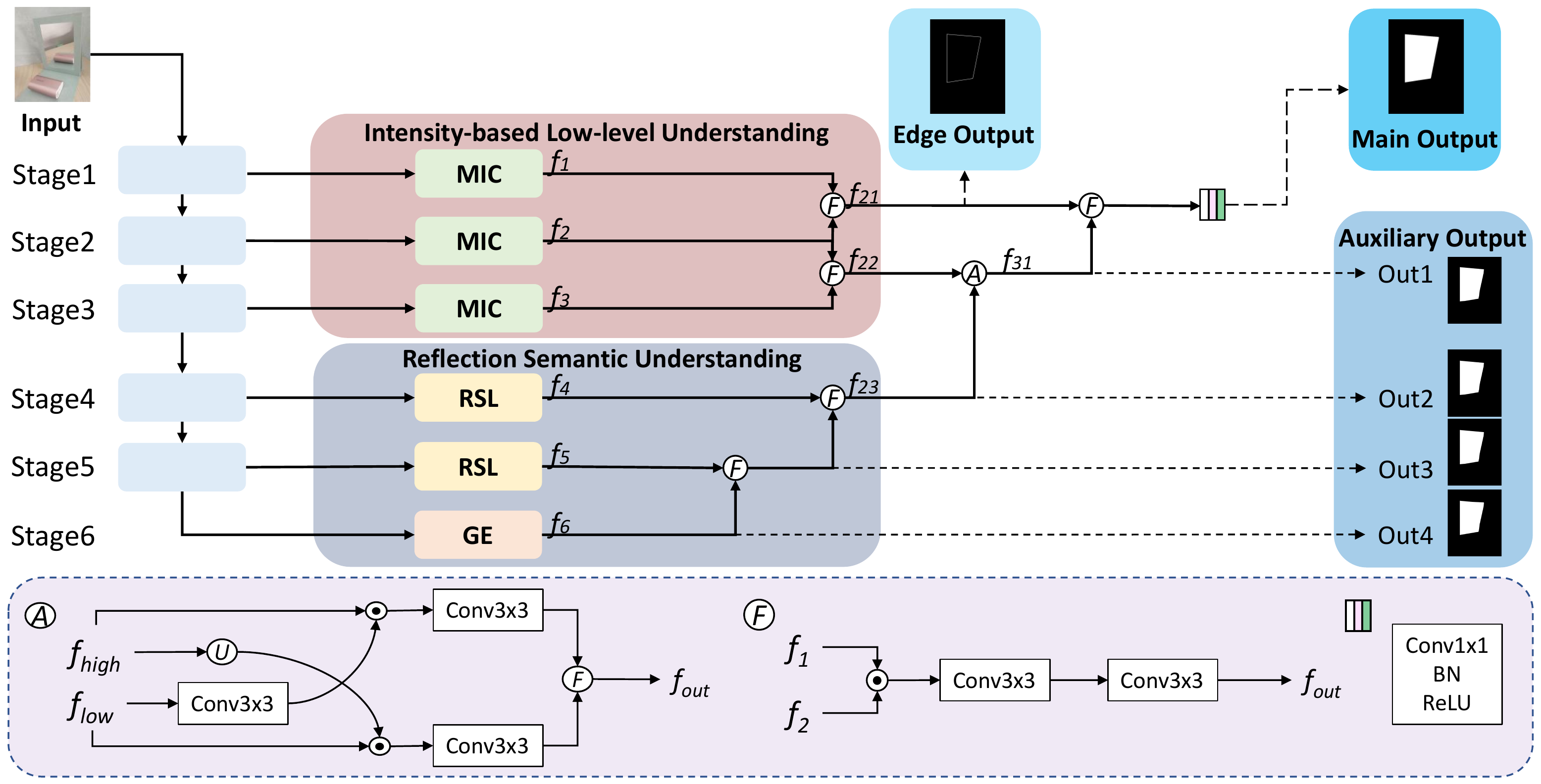}
    \caption{An overview of our proposed method. We first use ResNeXt-101 \cite{xie2017aggregated} as a backbone and a global extractor (GE) to extract multi-scale image features. We then apply multi-orientation intensity-based contrasted (MIC) modules at the first three stages and reflection semantic logical (RSL) modules at the remaining stages. We use edge and auxiliary output supervisions alongside the main output during the training process.}
    \label{fig:network}
\end{figure*}

HetNet is based on two observations. We observe that humans are easily attracted to regions with distinctive low-level features (\textit{e.g.}, intensity contrast) first, and then pay attention to high-level information (\textit{e.g.}, content similarity, contextual contrast) to check for object details to detect mirrors. These observations motivate us to learn low-level features at shallow stages and extract high-level features at deep stages with heterogeneous modules. Figure~\ref{fig:network} illustrates the pipeline.

\subsection{Overall Structure}
After obtaining the multi-scale image features from the backbone network~\cite{xie2017aggregated}, multi-orientation intensity-based contrasted (MIC) modules are used in the first three stages, while reflection semantic logical (RSL) modules and a global extractor (GE) are used in the last three, as shown in Figure~\ref{fig:network}. The global extractor (GE) extracts multi-scale image features following the pyramid pooling module~\cite{zhao2017pyramid}. Outputs after MIC, RSL and GE are denoted as $f_{i}$, where $i$ is the stage number starting from 1 to 6. Learning intensity-based low-level understandings, $f_{1}$ and $f_{2}$, $f_{2}$ and $f_{3}$ are then fused to $f_{21}$, $f_{22}$,
respectively. To integrate reflection semantic understandings, we fuse $f_{6}$ with $f_{5}$ first and then with $f_{4}$ to \ryn{produce} $f_{23}$. Finally, after $f_{23}$ is aggregated with $f_{22}$ to $f_{31}$, the output feature map is a fusion of $f_{31}$ and $f_{21}$. The fusion strategy is multiplication and two 3$\times$3 convolution layers with BatchNorm, 
where the low-level features $f_{low}$ and high-level features $f_{high}$ are fused in a cross aggregation strategy~\cite{zhao2021complementary,cai2020learning,chen2018cascaded}. First, the interim low-level features are computed from $f_{low}$ multiplies upsampled $f_{high}$.
The interim high-level features are the product of $f_{high}$ and $f_{low}$ processed by a 3$\times$3 convolution layer. The interim low-level and high-level features are fused after applying a 3$\times$3 convolution layer with BatchNorm and ReLU. 

\subsection{The MIC Module}

Only learning contextual information is insufficient, especially when contextual information is limited or complex. Thus, utilizing an additional strong cue to facilitate mirror detection is necessary. Gestalt psychology \cite{koffka2013principles} believes that most people see the whole scene first, and then pay attention to individual elements of the scene. In addition, the whole is not equivalent to the sum of the individual elements. Instead, it takes into account the degree of association of these elements (\textit{e.g.}, shape, position, size, color). Based on this, we believe that observing the same scene from different orientations may obtain different information. We use ICFEs to imitate orientation-selective preference visual cortex cells \cite{hubel1962receptive} to be proficient in learning features in one orientation. Strengthened contrast information is acquired by combining information learned from two single orientations. Considering low-level contrasts between mirror and non-mirror regions, we first design a MIC module to focus on two-orientation low-level contrasts to localize possible mirror regions.
\rz{In addition, to reduce computational costs, ICFEs process input features as 
two parallel 1D features in two directions separately. }

\begin{figure}[h]
    \centering
    \includegraphics[width=0.45\textwidth]{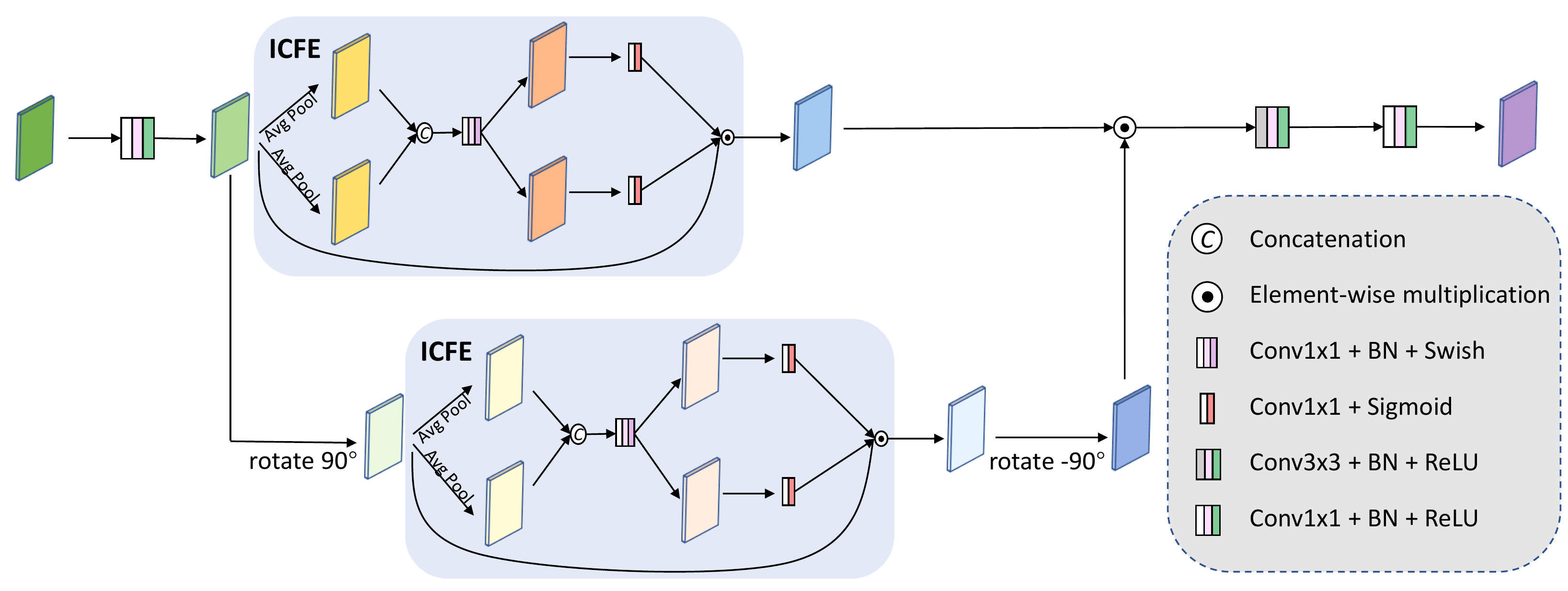}
    \caption{The architecture of the multi-orientation intensity-based contrasted (MIC) module. We first rotate the input features to obtain the other orientation and then use two Intensity-based Contrasted Feature Extractors (ICFEs) to extract low-level features in two orientations. After rotating the features back to the original orientation, we multiply them and take the product into convolution layers to compute low-level contrasts.}
    \label{fig:AM}
\end{figure}

A MIC module consists of two Intensity-based Contrasted Feature Extractor (ICFE) modules, as shown in Figure \ref{fig:AM}. Given the input image features after a 1$\times$1 convolution $f^{low}_{in}$, we first extract one orientation contrast features $f^{low}_{1}$ with ICFE directly and then extract contrast features $f^{low}_{2}$ at another orientation after rotating 90 degrees counterclockwise. 
To combine two-orientation contrasts into original orientation, we get $f^{low}_{3}$ by element-wise multiplication of $f^{low}_{1}$ and $f^{low}_{2}$ rotated back. Finally, we use a 3$\times$3 convolution and a 1$\times$1 convolution layer to extract the intensity-based contrasted low-level features $f^{low}_{out}$. Each of the convolution layers is followed by BatchNorm and ReLU.

\begin{align}
    & f^{low}_{1} = \textbf{ICFE}(f^{low}_{in}), \quad f^{low}_{2} = \textbf{ICFE}(\textit{Rot}(f^{low}_{in}, 1)),\\
    & f^{low}_{3} = f^{low}_{1} \odot \textit{Rot}(f^{low}_{2}, -1),\\
    & f^{low}_{out} =  \textbf{BConv}_{1 \times 1}(\textbf{BConv}_{3 \times 3}(f^{low}_{3})),
\end{align}

where \textit{Rot(f, d)} denotes rotating $f$ 90$^\circ$ for $d$ times.
$\odot$ represents element-wise multiplication. $\textbf{BConv}_{k \times k}(\cdot)$ refers to a \textit{k$\times$k} convolution with BatchNorm and ReLU activation function.

Inspired by the direction-aware strategy \cite{hou2021coordinate},
\rz{which embeds spatial information by a pair of 1D feature encoders instead of 2D global pooling,}
the input $f^{low}_{in}$ of ICFE first pools horizontally and vertically. The concatenated pooling results go through a 1$\times$1 convolution layer with BatchNorm and Swish function. After processing by two split branches, $f^{mid}_{h} \in \mathbb{R}^{C \times H \times 1}$  and $f^{mid}_{w} \in \mathbb{R}^{C \times 1 \times W}$ are applied a 1$\times$1 convolution layer and sigmoid function before multiplying together with $f^{low}_{in}$. Formally, we have:
\begin{align}
& f^{mid}_{1} = \textbf{SConv}_{1 \times 1}(\mathcal{P}_{h}(f^{low}_{in}) \copyright permute(\mathcal{P}_{v}(f^{low}_{in}))), \\
& f^{mid}_{h}, f^{mid}_{w} = split(f^{mid}_{1}), \\
& f^{mid}_{out} = \sigma(\textbf{Conv}_{1 \times 1}(f^{mid}_{h})) \odot \sigma(\textbf{Conv}_{1 \times 1}(f^{mid}_{w})) \odot f^{low}_{in},
\end{align}
where \textit{P$_{h, v}$($\cdot$)}, $\copyright$, $\sigma$ denote \textit{h}orizontal or \textit{v}ertical average pooling, concatenation, and Sigmoid, respectively. $\textbf{Conv}_{k \times k}(\cdot)$ represents a \textit{k$\times$k} convolution, and $\textbf{SConv}_{k \times k}(\cdot)$ refers to a $\textbf{Conv}_{k \times k}(\cdot)$ with BatchNorm and Swish activation function.

\subsection{The RSL Module}
As there are usually many low-level contrasts in the scene, it could be difficult to detect mirrors simply by low-level understandings. Owing to the reflection, parts outside the mirrors are similar to the contents inside mirrors. Besides, reflected contents may be distinctive to objects around mirrors, making content discontinuity a clue. Based on the above observations, we use a RSL module to learn high-level understandings to assist in finalizing mirror detection combined with the initial localization. 
\begin{figure}[h]
    \centering
    \includegraphics[width=0.5\textwidth]{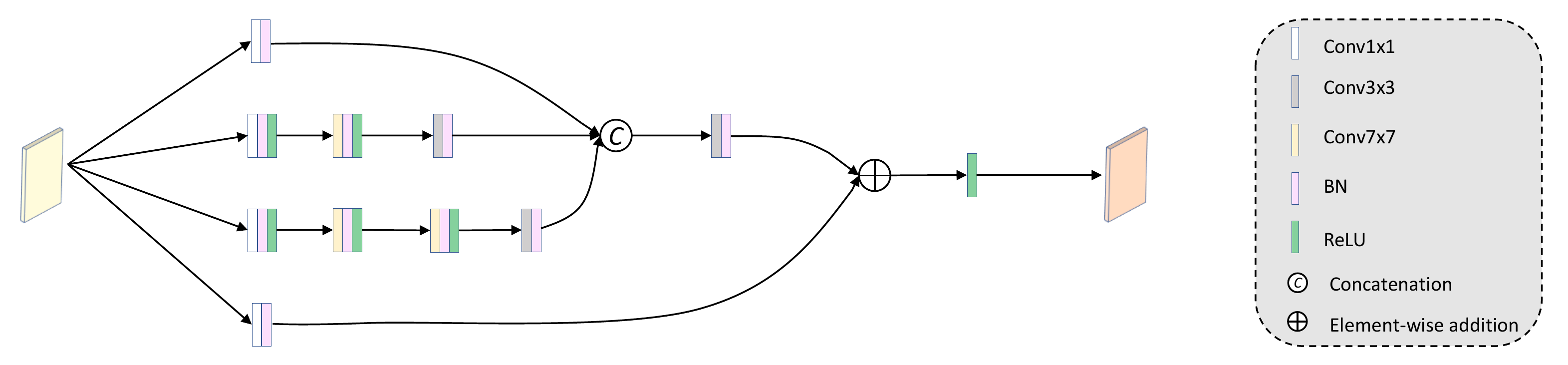}
    \caption{The architecture of the reflection semantic logical (RSL) module.}
    \label{fig:BM}
\end{figure}

RSL extracts information in four branches of different numbers of convolution layers with different kernel sizes and dilation rates.
Each branch extracts semantic features from a receptive field. Thus, integrating all branches expands a wider receptive field.
To obtain the output features $f^{s}_{i}$ of branch $i$ from the input high-level features $f^{high}_{in}$ in our RSL, we have:
\begin{align}
& f^{s}_{1} = \mathcal{N}(\textbf{Conv}_{1\times1}(f^{high}_{in})), \\
& f^{s}_{2} = \mathcal{N}(\textbf{Conv}_{3\times3}^{p,d=7}(\textbf{BConv}_{7\times7}^{p=3}(\textbf{BConv}_{1\times1}(f^{high}_{in})))), \\
& f^{s}_{3} = \mathcal{N}(\textbf{Conv}_{3\times3}^{p,d=7}(\textbf{BConv}_{7\times7}^{p=3}(\textbf{BConv}_{7\times7}^{p=3}(\textbf{BConv}_{1\times1}(f^{high}_{in}))))), \\
& f^{s}_{4} = \mathcal{N}(\textbf{Conv}_{1\times1}(f^{high}_{in})),
\end{align}
where \textit{N} denotes BatchNorm, and \textit{R} denotes ReLU. Superscripts \textit{p}, \textit{d} of \textbf{Conv}($\cdot$) related modules represent padding and dilation rate, respectively. The default padding and dilation rate are set as 1.

We then combine the output features from all branches:
\begin{align}
& f^{s}_{mid} = \mathcal{N}(\textbf{Conv}_{3\times3}(f^{s}_{1} \copyright f^{s}_{2} \copyright f^{s}_{3})), \\
& f^{s}_{out} = \mathcal{R}( f^{s}_{mid}+f^{s}_{4}).
\end{align}

Under such design, our RSL can acquire rich reflection semantic logical information to determine real mirrors from potential regions predicted by the previous modules.

\subsection{Loss Function}
During the training process, we apply multi-scale supervision for mirror maps and also supervise edge extraction for initial localization. We use the pixel position aware (PPA) loss \cite{wei2020f3net} for multi-scale mirror map supervision, and binary cross entropy (BCE) loss for mirror edge supervision. The PPA loss is the sum of weighted BCE (wBCE) loss and weighted IoU (wIoU) loss. wBCE loss concentrates more on hard pixels (\textit{e.g.}, holes) than the BCE loss \cite{de2005tutorial}. wIoU measures global structure and pays more attention to important pixels than the IoU loss \cite{mattyus2017deeproadmapper}. The final loss function is therefore:
\begin{align}
    Loss = L_{bce} + \sum\limits^{4}_{i=0}\frac{1}{2^{i}}L_{ppa}^{i},
\end{align}
where $L_{ppa}$ is the pixel position aware (PPA) loss between the i-th mirror map and the ground truth mirror map, while $L_{bce}$ is the binary cross-entropy (BCE) loss.

\section{Experiments}
\subsection{Datasets}
We conduct experiments on two datasets: MSD \cite{yang2019my} and PMD \cite{lin2020progressive}. MSD focuses more on similar indoor scenes, but PMD contains diverse scenes. MSD includes 3,063 images for training and 955 for testing, while PMD has 5,096 images for training and 571 for testing. We train our method on each training set and then test it separately.

\subsection{Evaluation Metrics}
We adopt three evaluation metrics: Mean Absolute Error (MAE), Intersection over union (IoU), and F-measure to evaluate the performances of the models quantitatively.
MAE represents average pixel-wise error between the prediction mask and ground truth.
F-measure (F$_{\beta}$) is a trade-off between precision and recall. It is computed as:
\[F_{\beta} = \frac{(1 + \beta^{2}) \cdot Precision \cdot Recall}{\beta^{2} \cdot Precision + Recall},\]
where $\beta^{2}$ is set to 0.3, for precision is more important \cite{achanta2009frequency}. Larger F$_{\beta}$ is better.

\subsection{Implementation Details}
We implement our model by PyTorch and conduct experiments on a GeForce RTX2080Ti GPU. We use ResNeXt-101 \cite{xie2017aggregated} pretrained on ImageNet as our backbone network. Input images are resized to multi-scales with random crops and horizontal flips during the training process. We use the stochastic gradient descent (SGD) optimizer with a momentum value of 0.9 and a weight decay of 5e-4. In the training phase, the maximum learning rate is 1e-2, the batch size is 12, and the training epoch is 150. It takes around 5 hours to train. As for the inference process, input images are only resized to 352$\times$352 and then directly predict final maps without any post-processing.

\subsection{Comparison to the State-of-the-art Methods}
To prove the effectiveness and efficiency of our method, we compare it with 10 state-of-the-art methods, including salient object detection methods (R$^{3}$Net \cite{deng2018r3net}, CPDNet \cite{wu2019cascaded}, EGNet \cite{zhao2019egnet}, LDF \cite{wei2020label}, MINet \cite{pang2020multi}, SETR \cite{zheng2021rethinking}, VST \cite{liu2021visual}), and mirror detection methods MirrorNet\cite{yang2019my}, PMD\cite{lin2020progressive}, SANet\cite{guan2022learning}.
Table~\ref{tab:eval} illustrates the quantitative comparison on the three metrics. Our method achieves the best performances on all metrics. Besides, as shown in Table~\ref{tab:efficiency}, we compare Parameters, FLOPs, and FPS with relevant methods. 
As the hardware environment influences FPS, we conduct all the experiments on the same PC to ensure fairness. Quantitative results show that our method meets the balance between efficiency and accuracy.

\begin{table}[t]
\centering
\caption{Quantitative comparison with the state-of-the-art methods on two benchmarks with evaluation metrics MAE, IoU, and F$_{\beta}$. The best  results are shown in \textcolor{red}{red} and the second best results are shown in \textcolor{blue}{blue}.}
\scalebox{0.8}{
\begin{tabular}{l|lll|lll}
\hline
          & \multicolumn{3}{c|}{MSD}                                                    & \multicolumn{3}{c}{PMD}                                                    \\ \cline{2-7} 
Method    & \multicolumn{1}{c}{MAE$\downarrow$} & \multicolumn{1}{c}{IoU$\uparrow$} & \multicolumn{1}{c|}{F$_{\beta}\uparrow$} & \multicolumn{1}{c}{MAE$\downarrow$} & \multicolumn{1}{c}{IoU$\uparrow$} & \multicolumn{1}{c}{F$_{\beta}\uparrow$} \\ \hline
R$^{3}$Net     & 0.111                   & 0.554                   & 0.767                   & 0.045                   & 0.496                   & 0.713                  \\
CPDNet    & 0.116                   & 0.576                   & 0.743                   & 0.041                   & 0.600                   & 0.734                  \\
EGNet     & 0.096                   & 0.630                   & 0.779                   & 0.088                   & 0.210                   & 0.590                  \\
LDF       & 0.068                   & 0.729                   & 0.843                   & 0.038                   & 0.633                   & 0.783                  \\
MINet     & 0.088                   & 0.664                   & 0.817                   & 0.038                   & 0.608                   & 0.765                  \\
SETR      & 0.071                   & 0.690                   & 0.851                   & 0.035                   & 0.564                   & \textcolor{blue}{0.797}                  \\
VST       & 0.054                   & 0.791                   & 0.871                   & 0.036                   & 0.591                   & 0.736                  \\ \hline
MirrorNet & 0.065                   & 0.790                   & 0.857                   & 0.043                   & 0.585                   & 0.741                  \\
PMD       & \textcolor{blue}{0.047}                   & \textcolor{blue}{0.815}                   & \textcolor{blue}{0.892 }                  & \textcolor{blue}{0.032}                   & 0.660               & 0.794                  \\ 
SANet &0.054 	&0.798 &	0.877 	&\textcolor{blue}{0.032} 	&\textcolor{blue}{0.668} 	&0.795 
\\ \hline
HetNet    & \textcolor{red}{0.043}                   & \textcolor{red}{0.828}                   & \textcolor{red}{0.906}                   & \textcolor{red}{0.029}                   & \textcolor{red}{0.690}                   & \textcolor{red}{0.814}                  \\ \hline
\end{tabular}
}
\label{tab:eval}
\end{table}

\begin{table}[t]
\centering
\caption{Quantitative comparison on efficiency. We compare our model with relevant state-of-the-art models on Parameters(M), FLOPs(GMAC), and FPS.}
\scalebox{0.8}{
\begin{tabular}{l|c|ccc}
\hline
Method    & Input Size     & Para. & FLOPs & FPS    \\ \hline
R$^{3}$Net     & 300$\times$300 & 56.16         & 47.53        & 8.30   \\
CPDNet    & 352$\times$352 & 47.85         & 17.77        & 65.83  \\
EGNet     & 256$\times$256 & 111.64        & 157.21       & 10.76  \\
LDF       & 352$\times$352 & 25.15         & 15.51        & 107.74 \\
MINet     & 320$\times$320 & 162.38        & 87.11        & 3.55   \\
SETR      & 480$\times$480 & 91.76         & 85.41        & 2.84   \\
VST       & 224$\times$224 & 44.48         & 23.18        & 6.05   \\ \hline
MirrorNet & 384$\times$384 & 121.77        & 77.73        & 7.82   \\
PMD       & 384$\times$384 & 147.66        & 101.54       & 7.41   \\ 
SANet     & 384$\times$384 & 104.80 	   & 66.00 	      & 8.53   \\ \hline
HetNet    & 352$\times$352 & 49.92         & 27.69        & 49.23  \\ \hline
\end{tabular}
}
\label{tab:efficiency}
\end{table}

\begin{figure*}[t] \centering
    \includegraphics[width=0.06\textwidth]{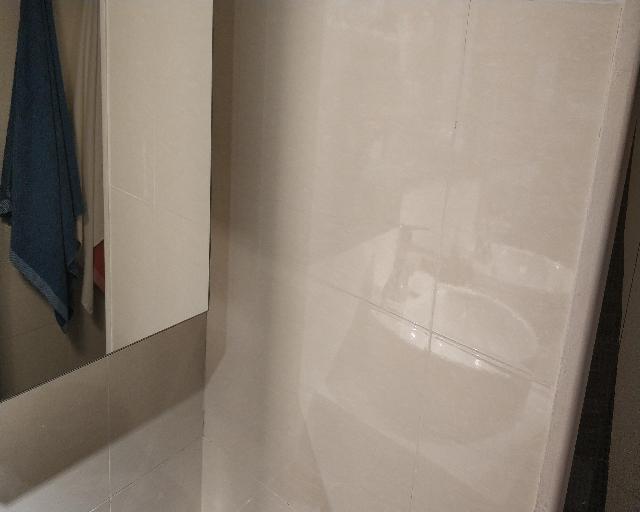}
    \includegraphics[width=0.06\textwidth]{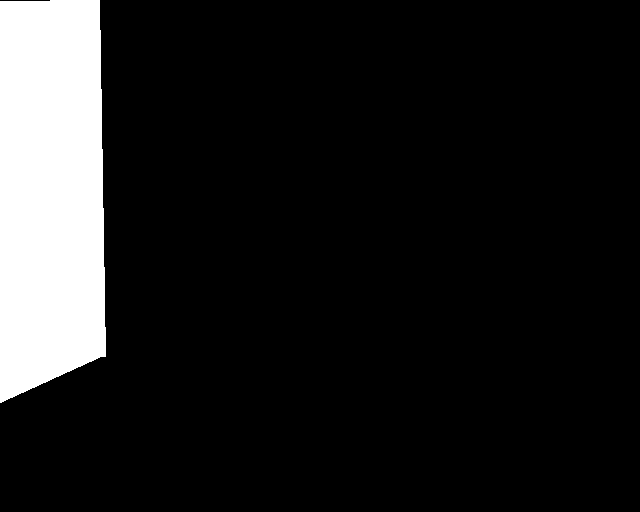}
    \includegraphics[width=0.06\textwidth]{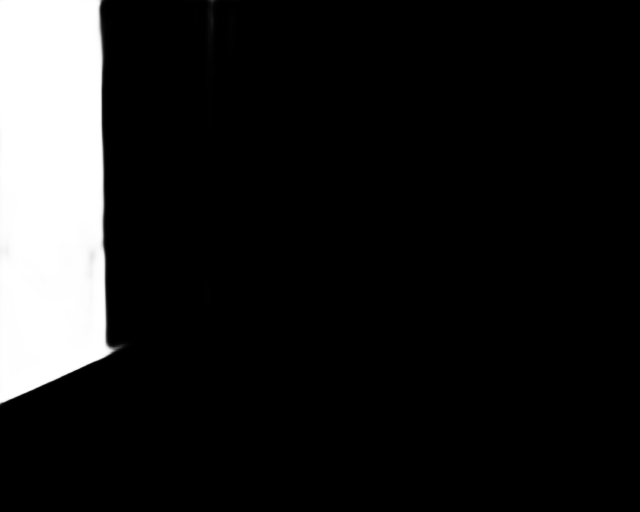}
    \includegraphics[width=0.06\textwidth]{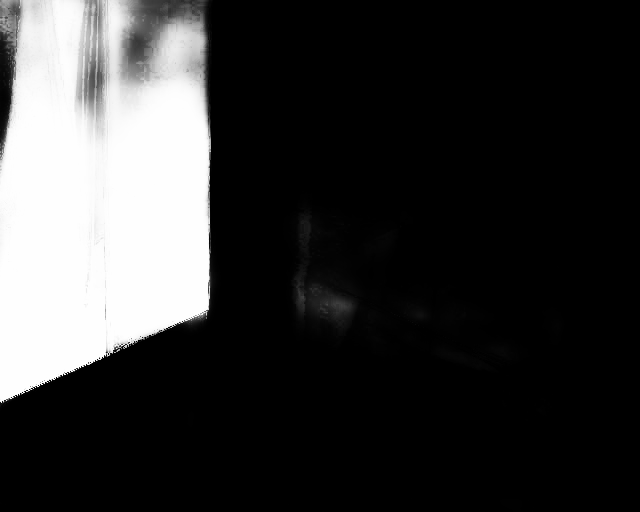}
    \includegraphics[width=0.06\textwidth]{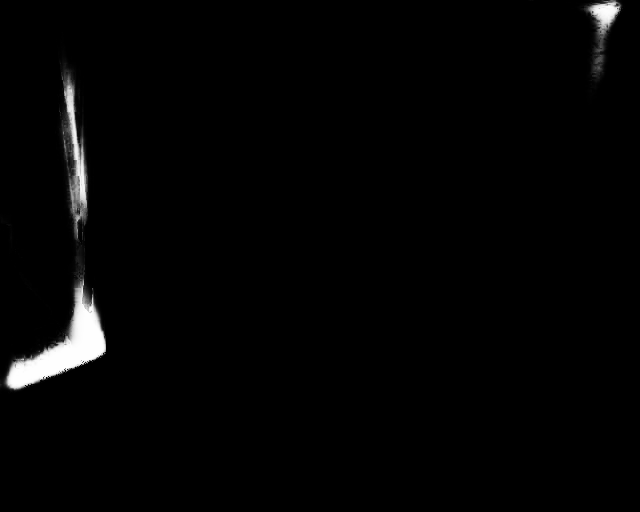}
    \includegraphics[width=0.06\textwidth]{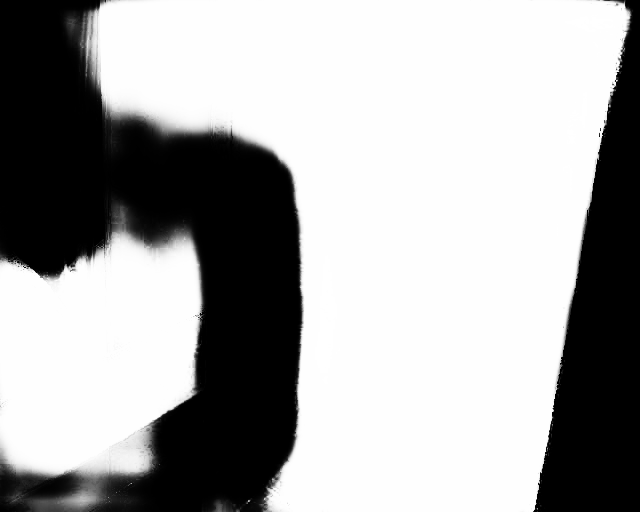}
    \includegraphics[width=0.06\textwidth]{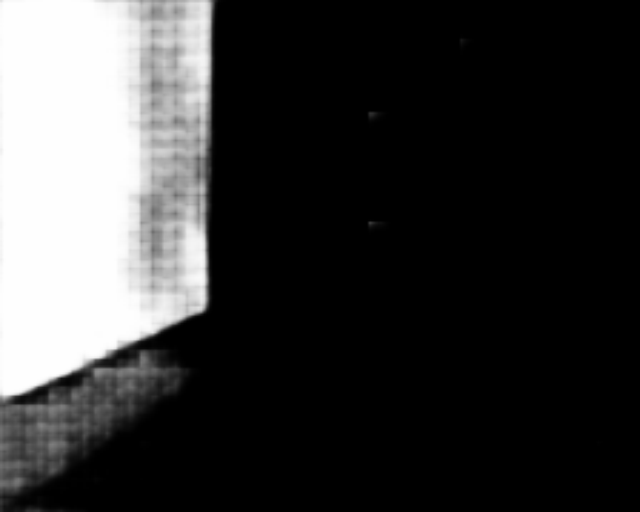}
    \includegraphics[width=0.06\textwidth]{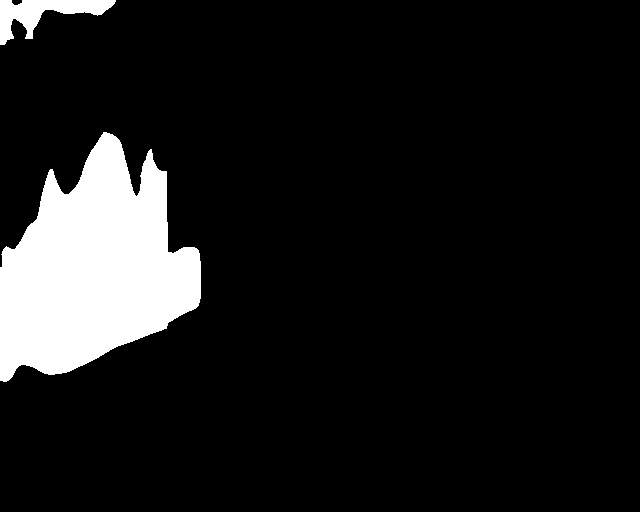}
    \includegraphics[width=0.06\textwidth]{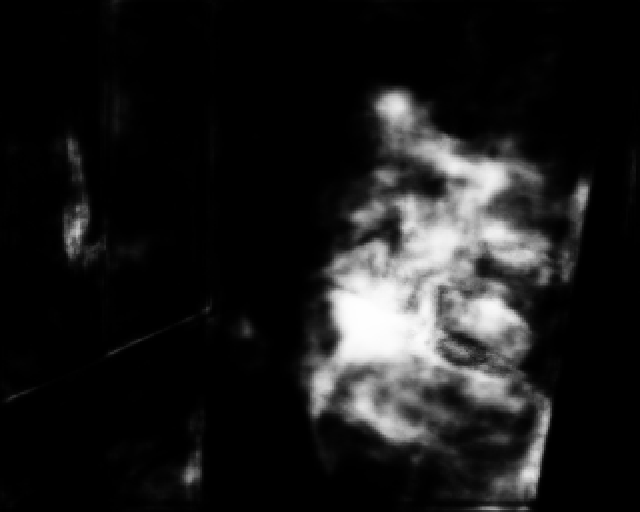}
    \includegraphics[width=0.06\textwidth]{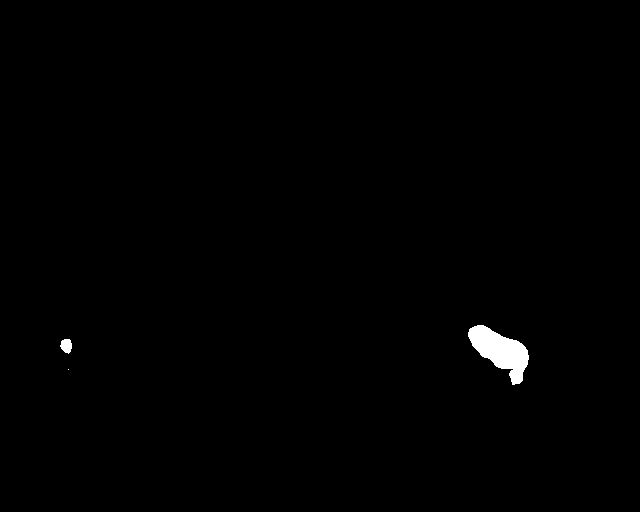}
    \includegraphics[width=0.06\textwidth]{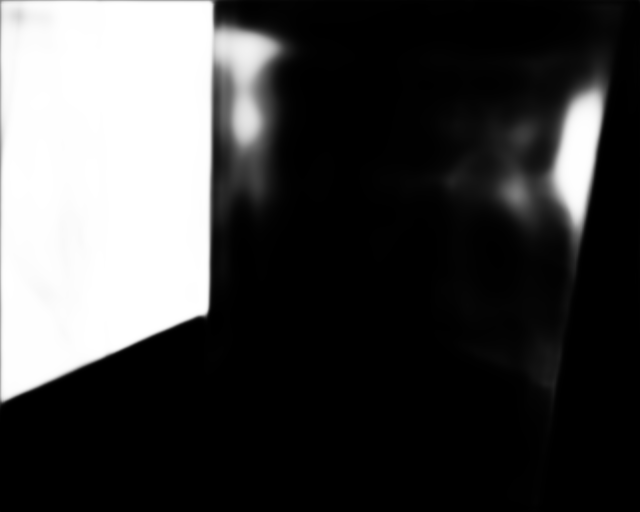}
    \includegraphics[width=0.06\textwidth]{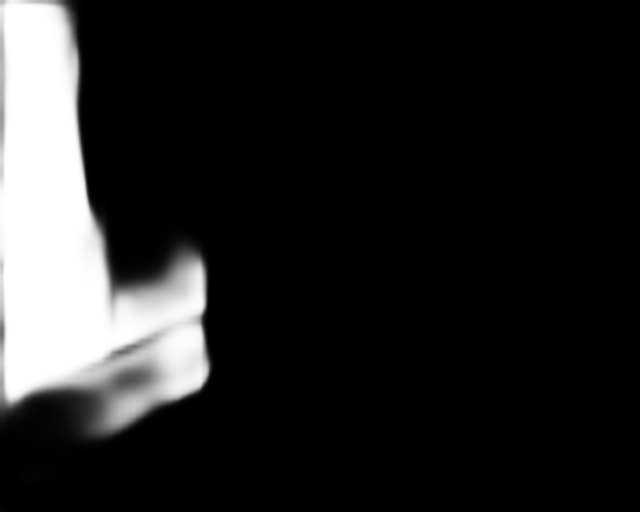}
    \includegraphics[width=0.06\textwidth]{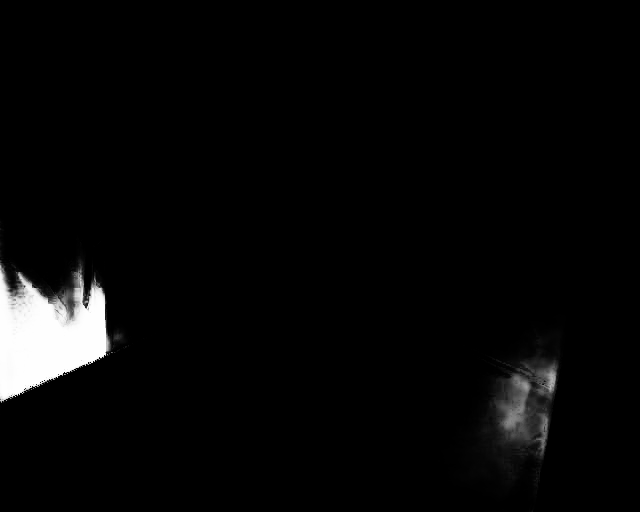}
    \\
    \includegraphics[width=0.06\textwidth]{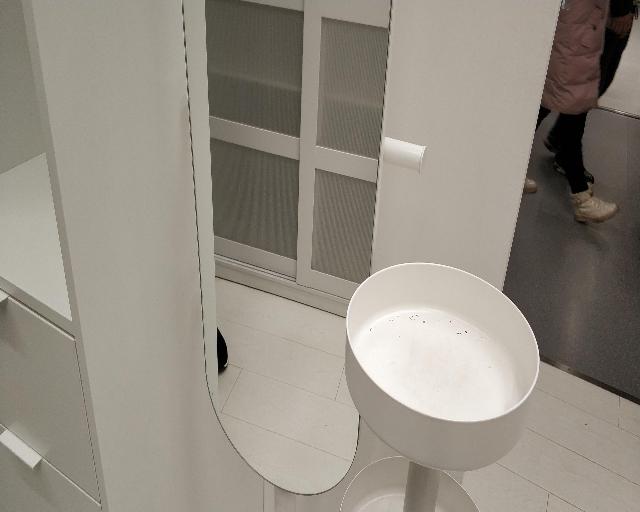}
    \includegraphics[width=0.06\textwidth]{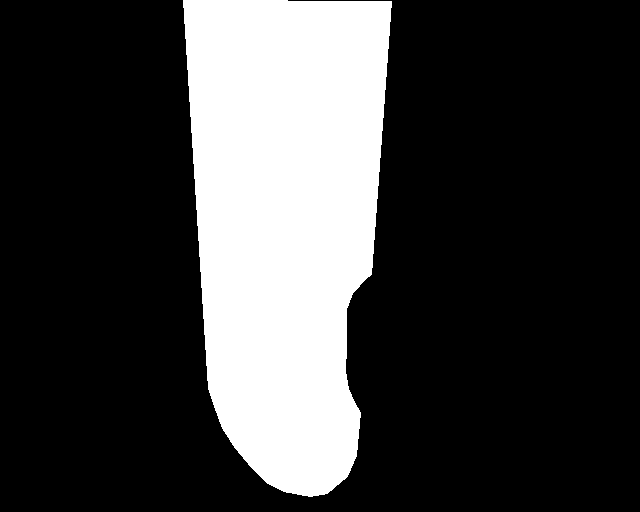}
    \includegraphics[width=0.06\textwidth]{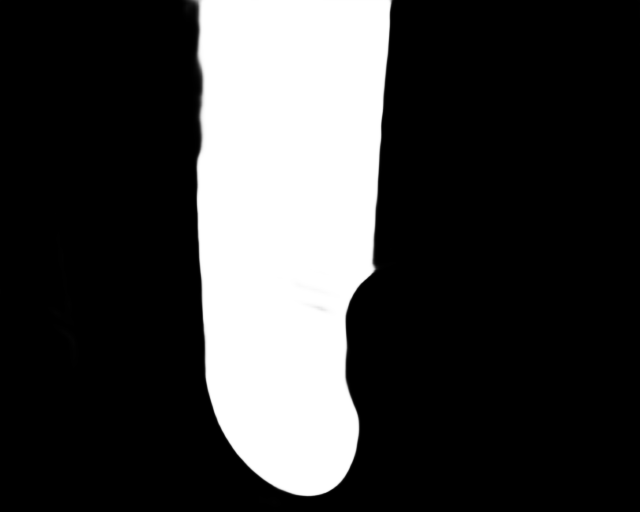}
    \includegraphics[width=0.06\textwidth]{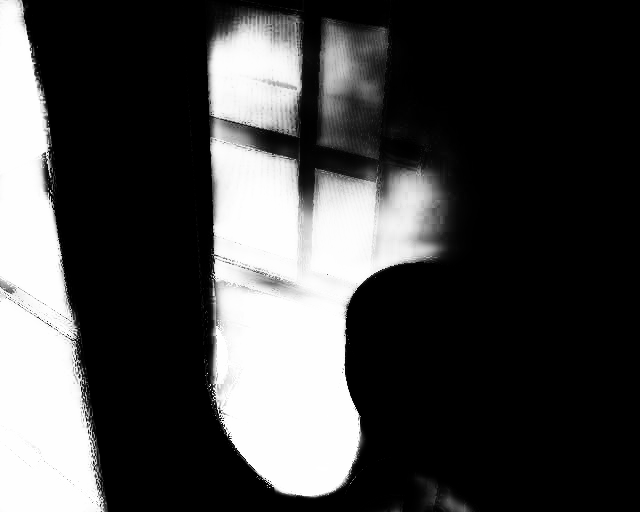}
    \includegraphics[width=0.06\textwidth]{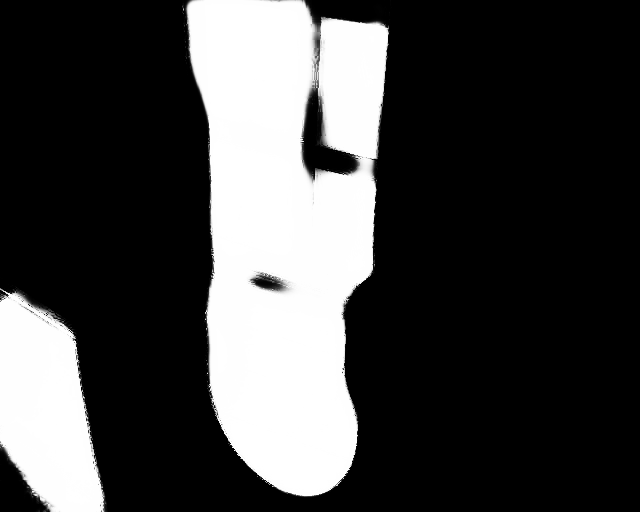}
    \includegraphics[width=0.06\textwidth]{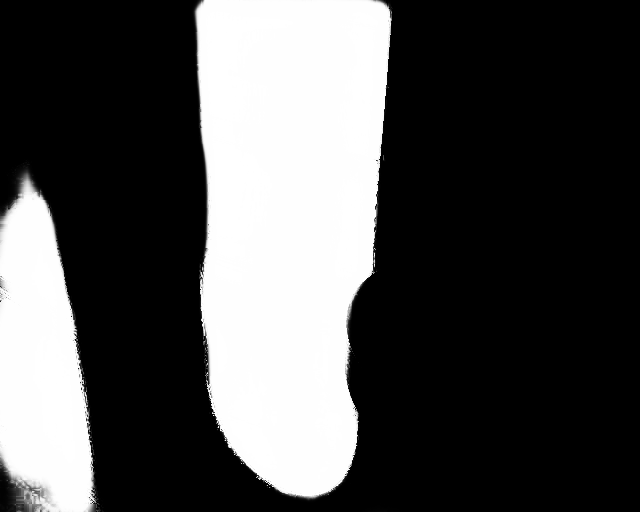}
    \includegraphics[width=0.06\textwidth]{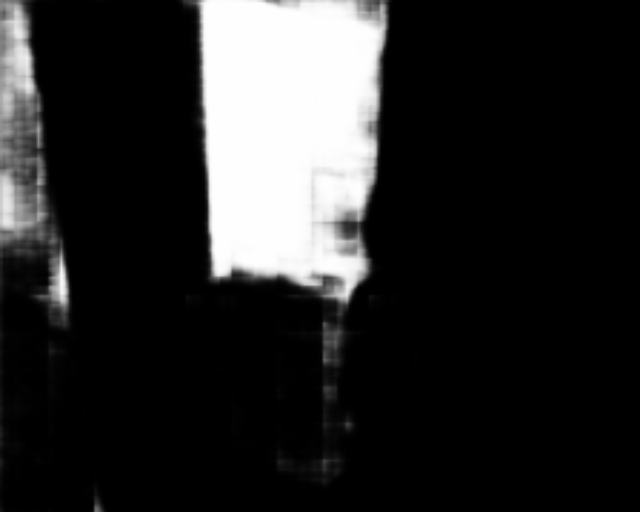}
    \includegraphics[width=0.06\textwidth]{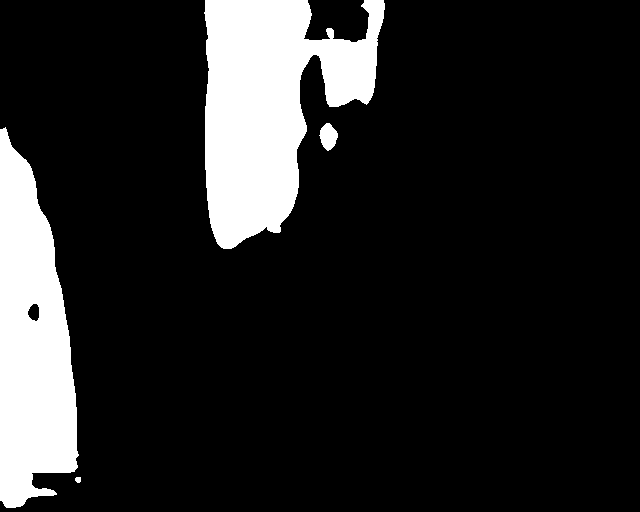}
    \includegraphics[width=0.06\textwidth]{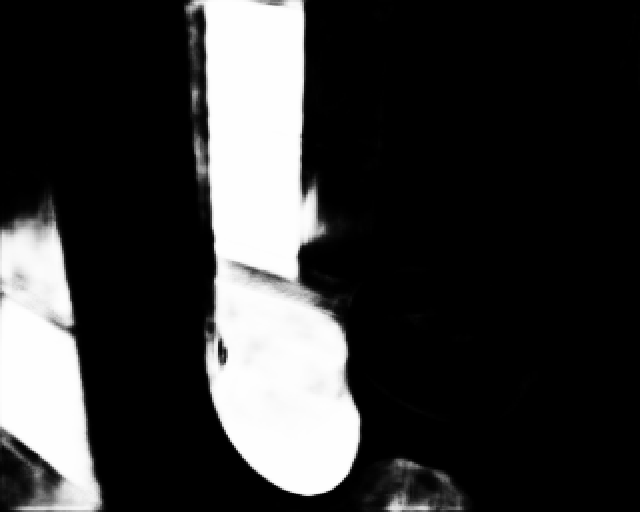}
    \includegraphics[width=0.06\textwidth]{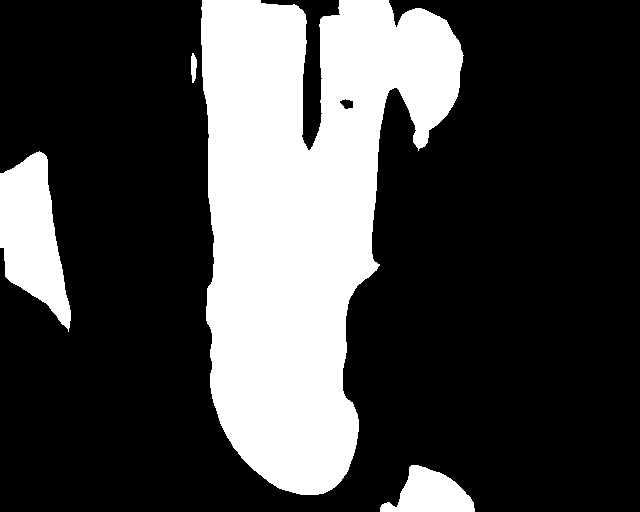}
    \includegraphics[width=0.06\textwidth]{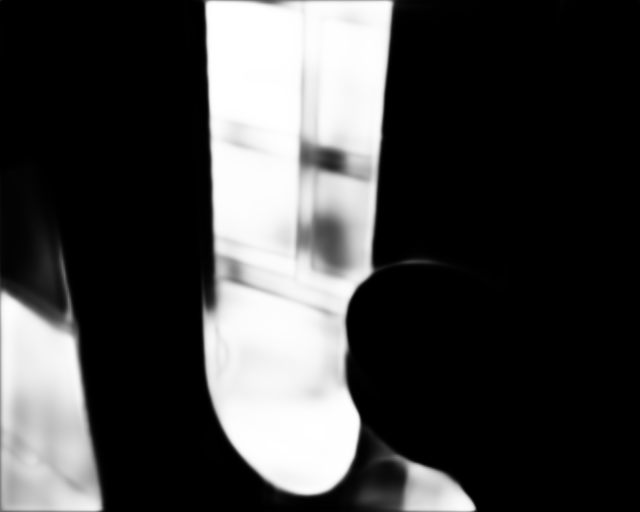}
    \includegraphics[width=0.06\textwidth]{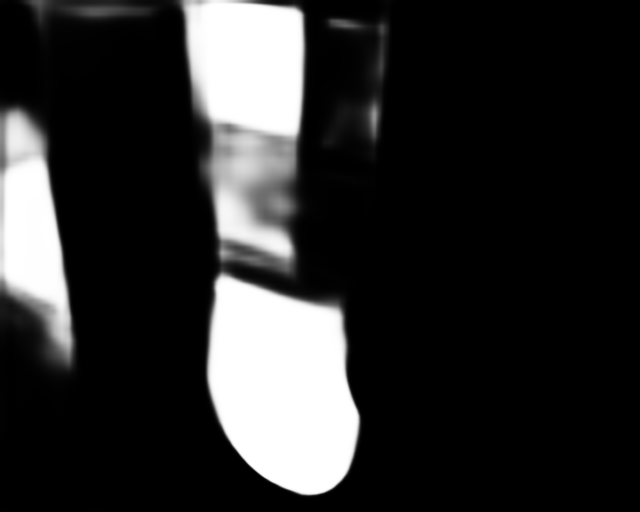}
    \includegraphics[width=0.06\textwidth]{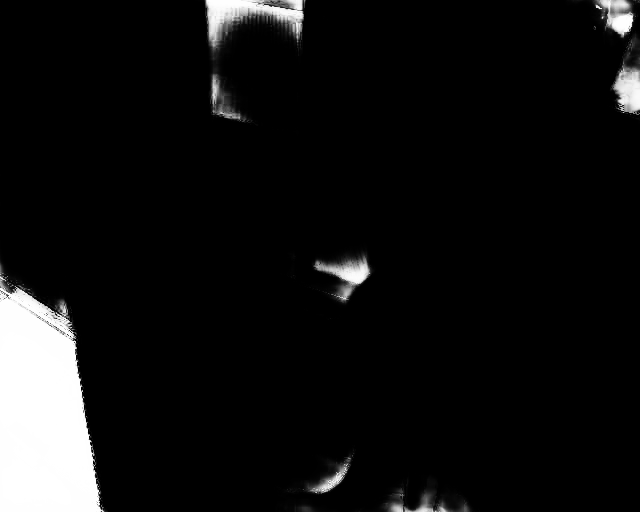}  
    \\
    \includegraphics[width=0.06\textwidth]{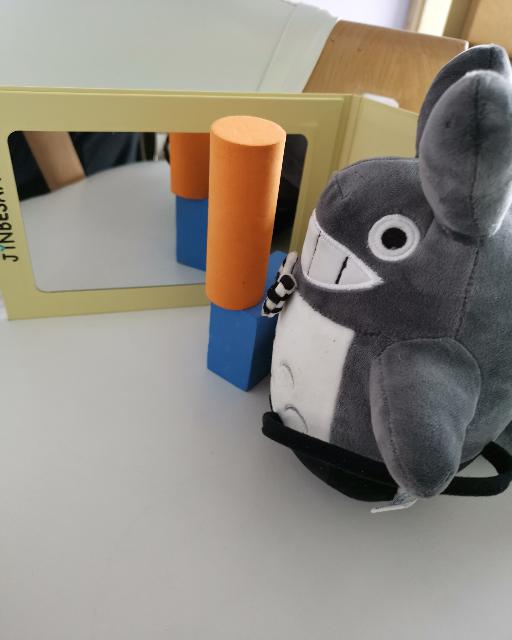}
    \includegraphics[width=0.06\textwidth]{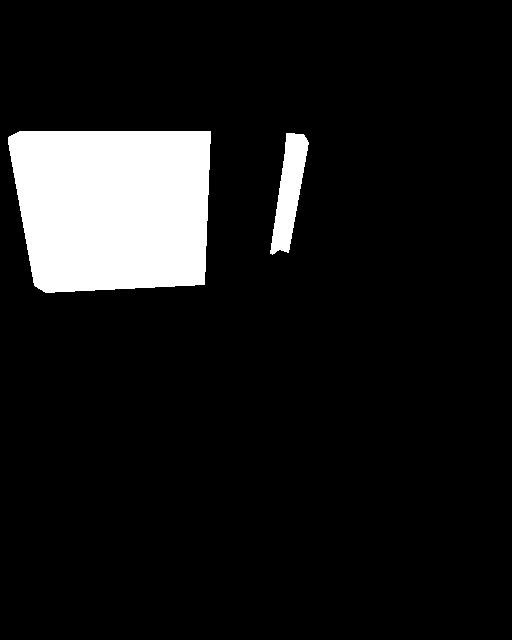}
    \includegraphics[width=0.06\textwidth]{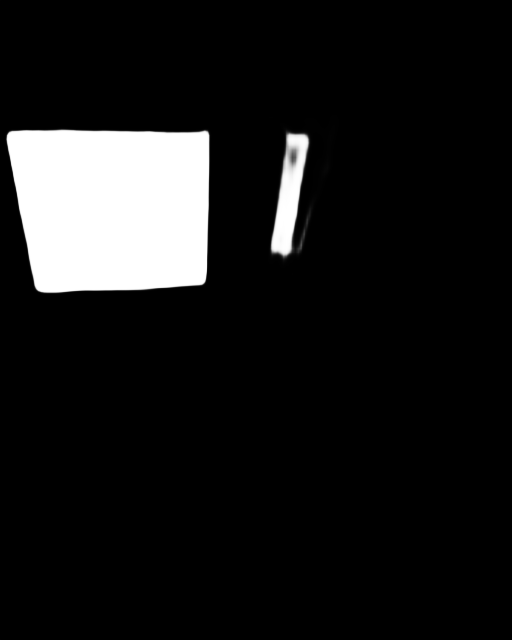}
    \includegraphics[width=0.06\textwidth]{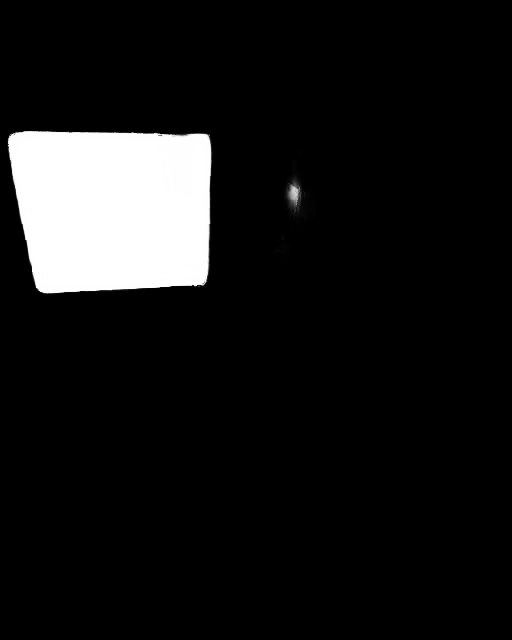}
    \includegraphics[width=0.06\textwidth]{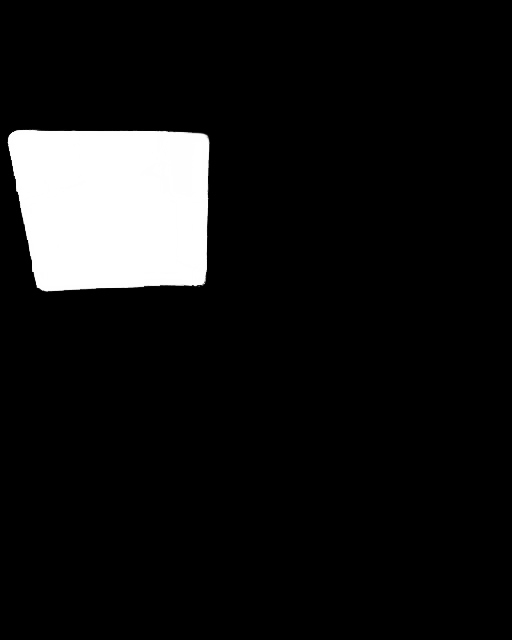}
    \includegraphics[width=0.06\textwidth]{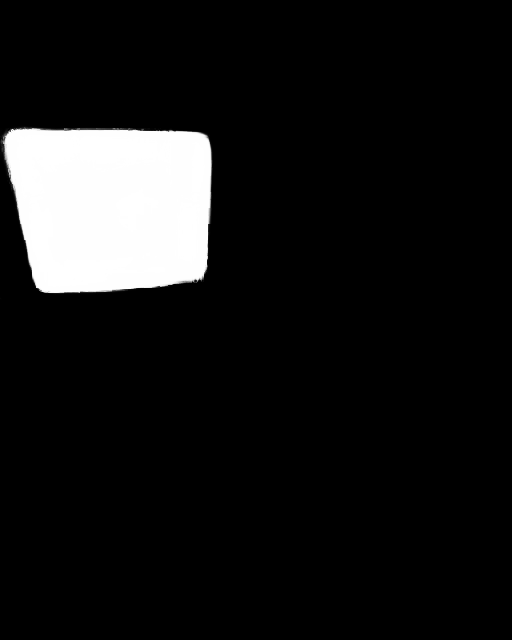}
    \includegraphics[width=0.06\textwidth]{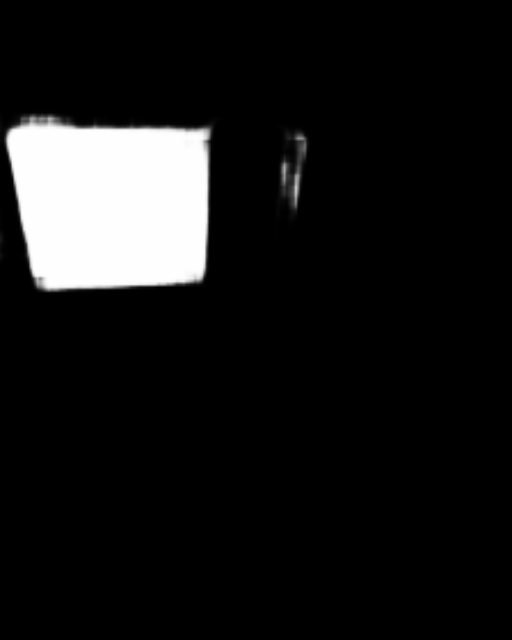}
    \includegraphics[width=0.06\textwidth]{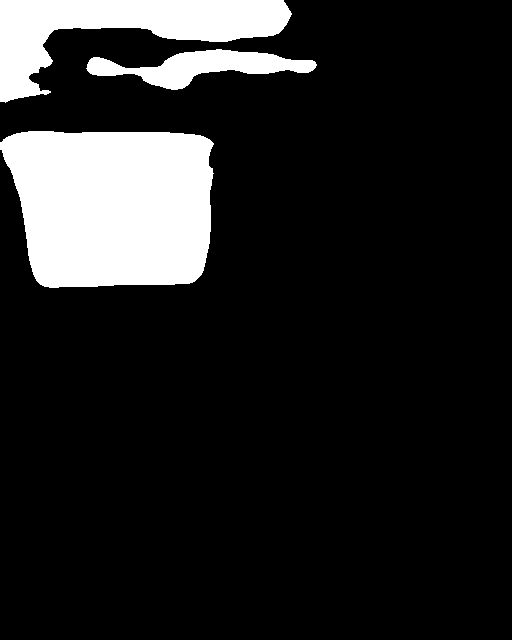}
    \includegraphics[width=0.06\textwidth]{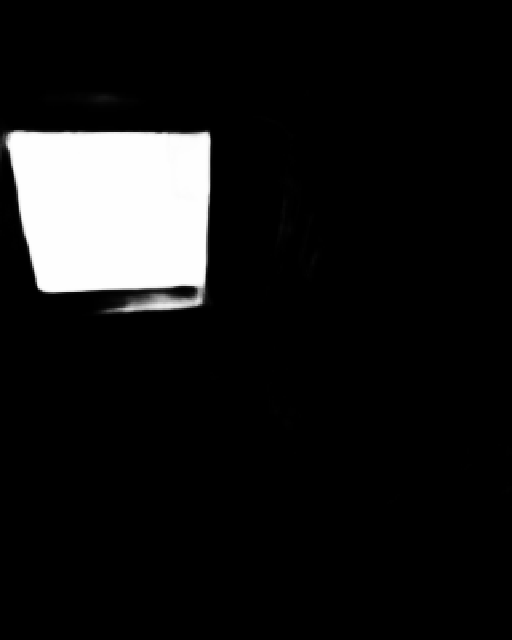}
    \includegraphics[width=0.06\textwidth]{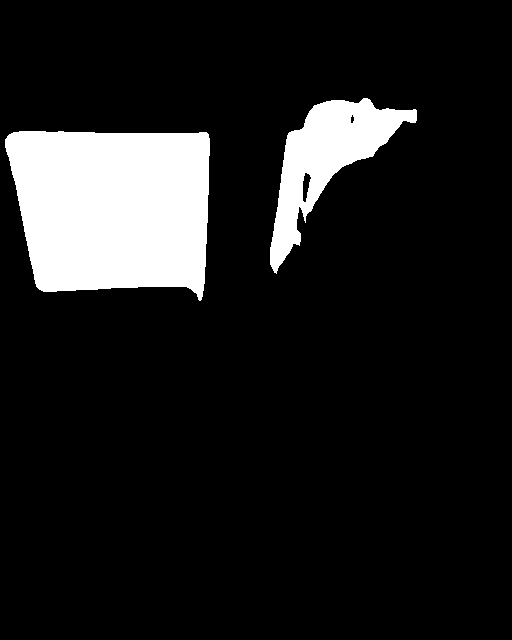}
    \includegraphics[width=0.06\textwidth]{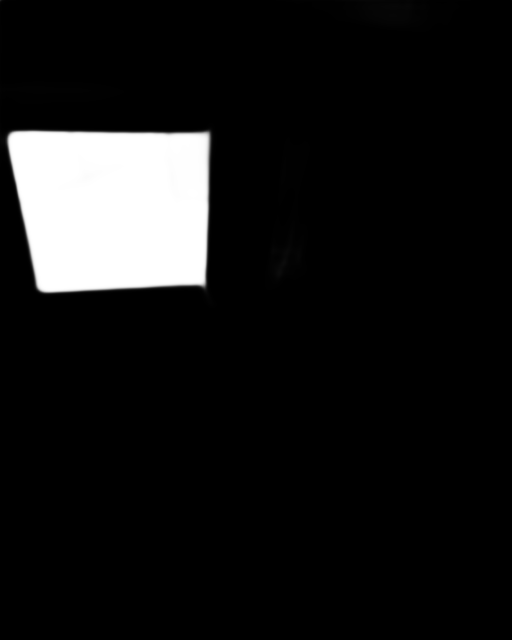}
    \includegraphics[width=0.06\textwidth]{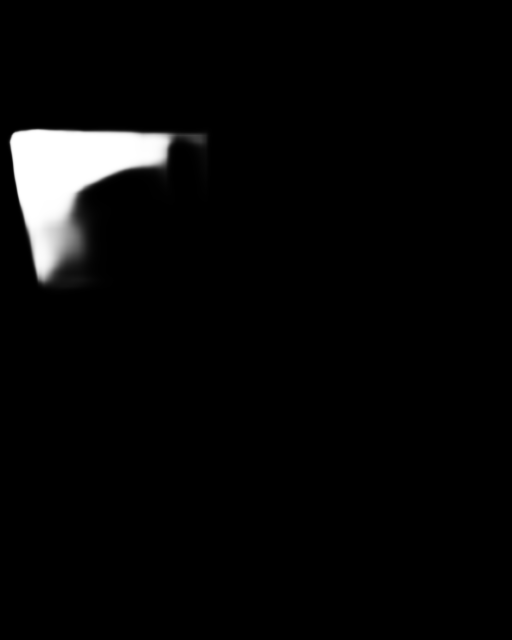}
    \includegraphics[width=0.06\textwidth]{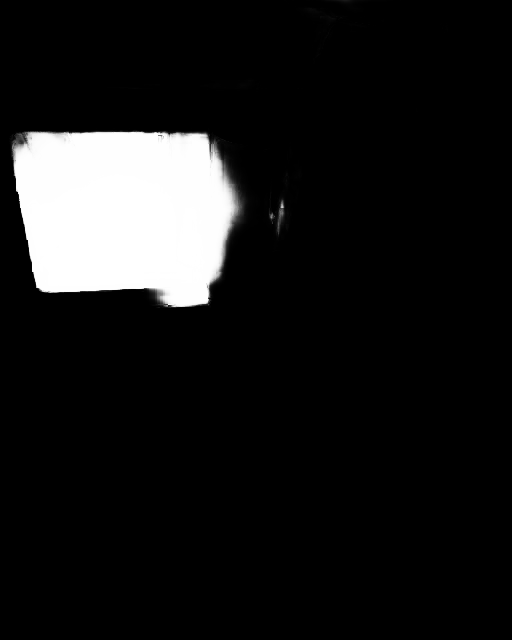}    
    \\
    \includegraphics[width=0.06\textwidth]{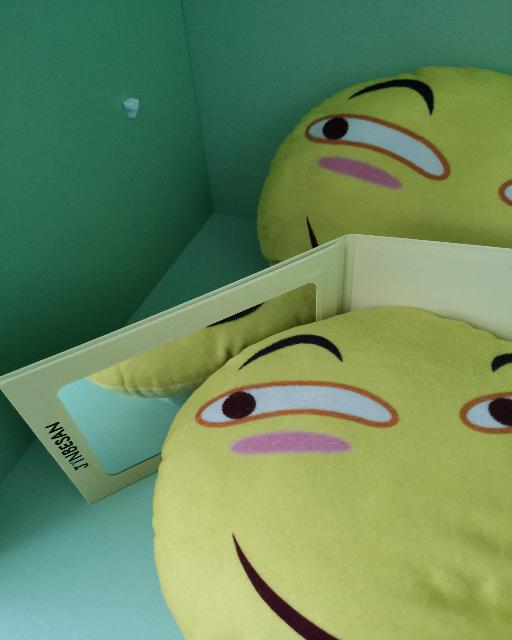}
    \includegraphics[width=0.06\textwidth]{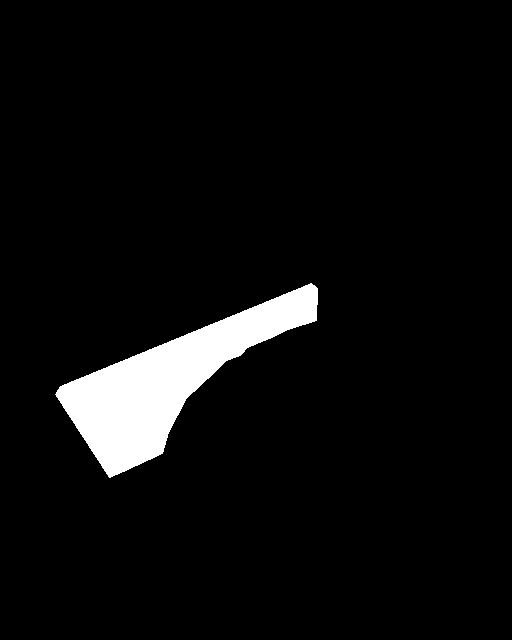}
    \includegraphics[width=0.06\textwidth]{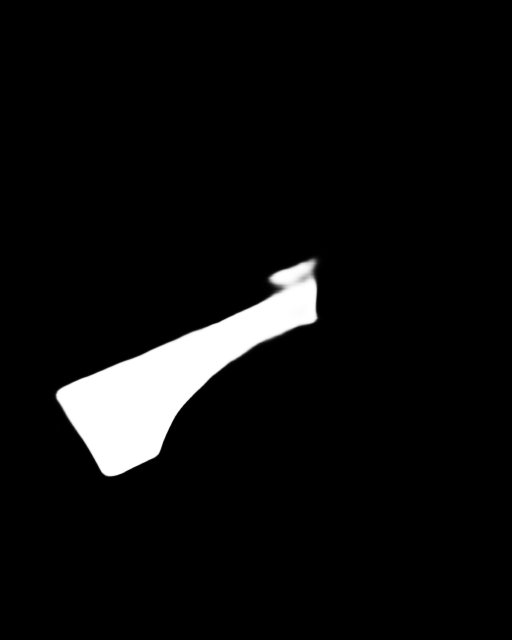}
    \includegraphics[width=0.06\textwidth]{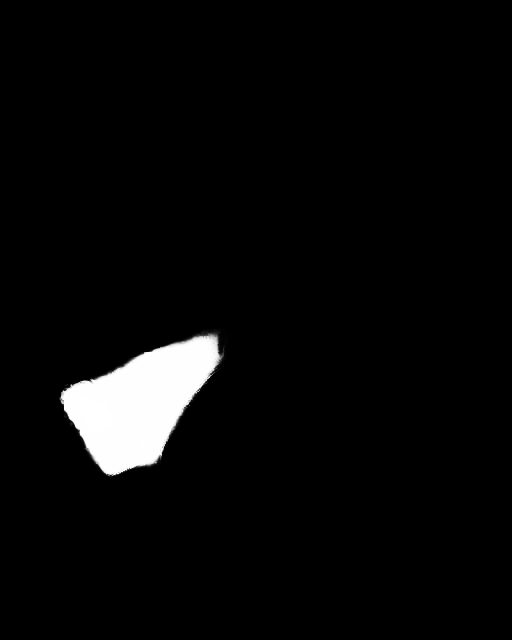}
    \includegraphics[width=0.06\textwidth]{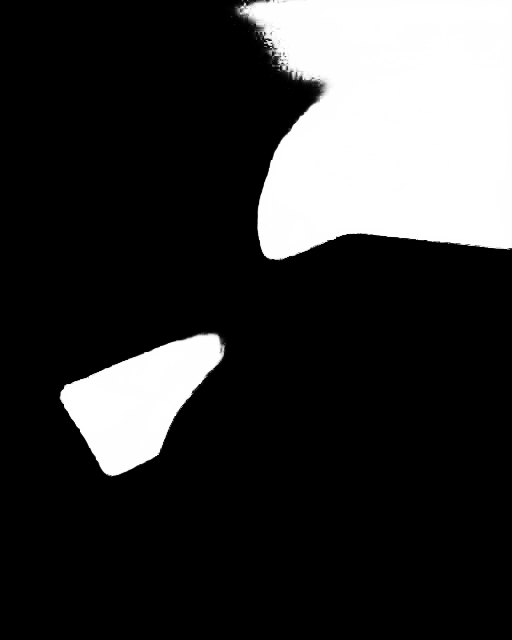}
    \includegraphics[width=0.06\textwidth]{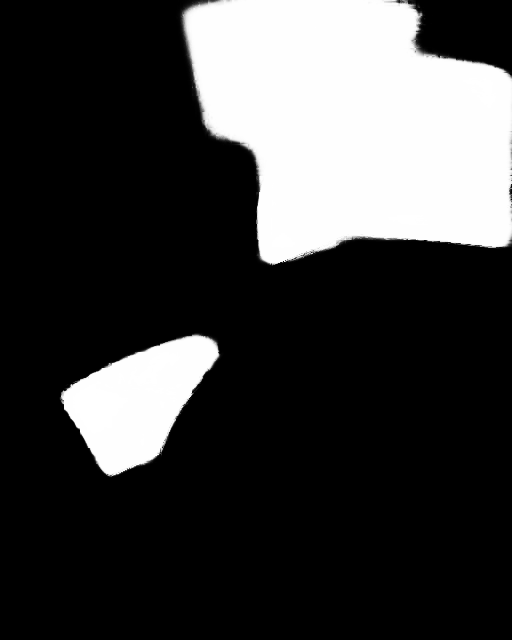}
    \includegraphics[width=0.06\textwidth]{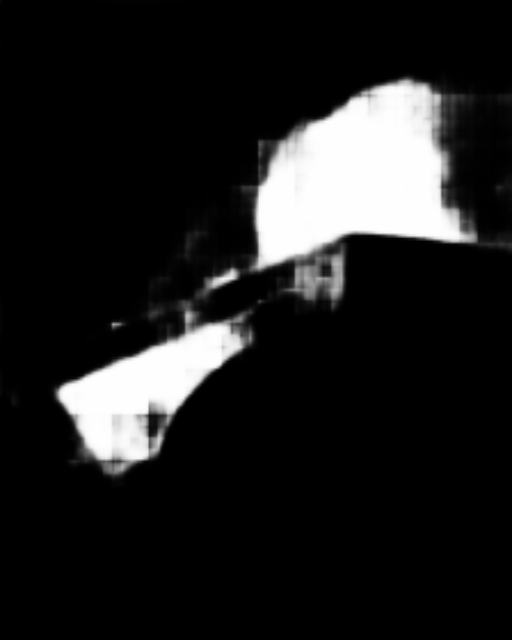}
    \includegraphics[width=0.06\textwidth]{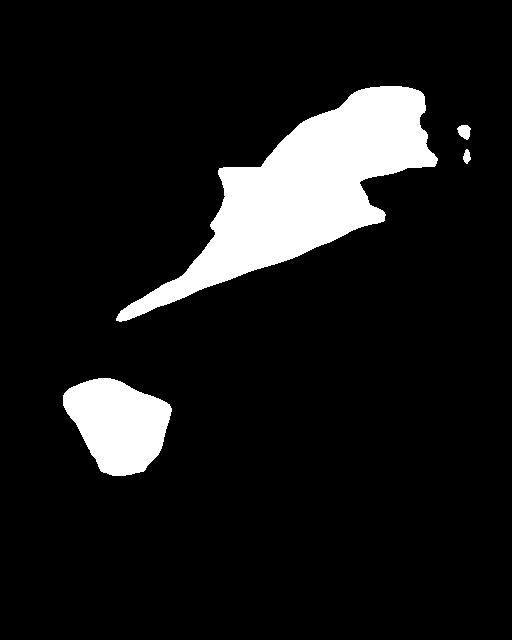}
    \includegraphics[width=0.06\textwidth]{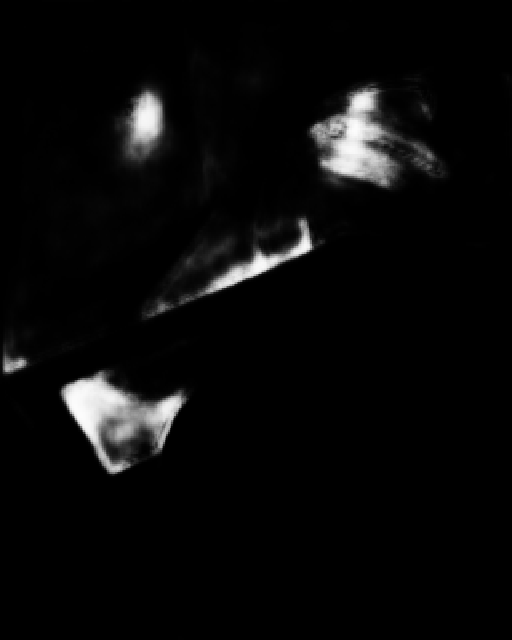}
    \includegraphics[width=0.06\textwidth]{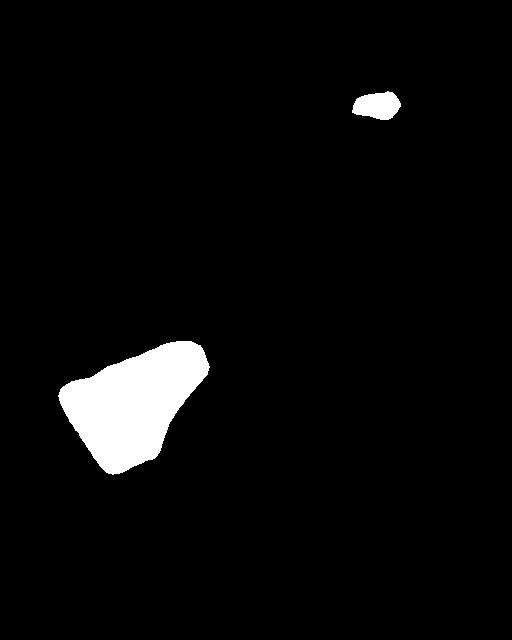}
    \includegraphics[width=0.06\textwidth]{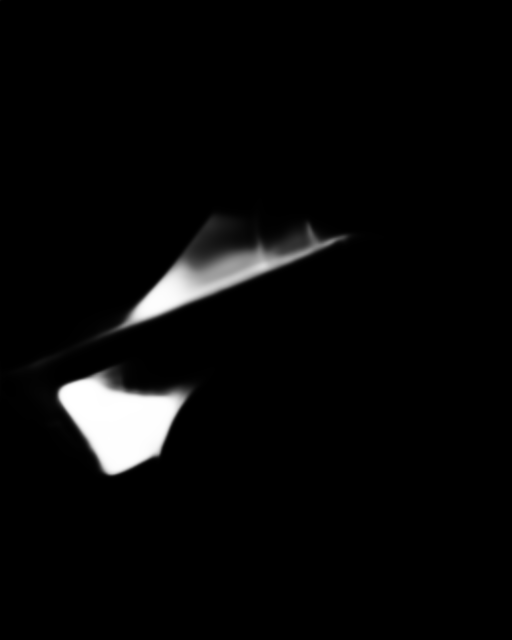}
    \includegraphics[width=0.06\textwidth]{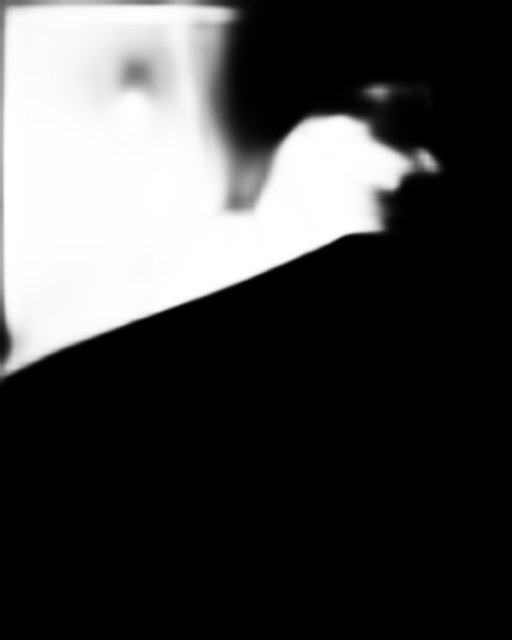}
    \includegraphics[width=0.06\textwidth]{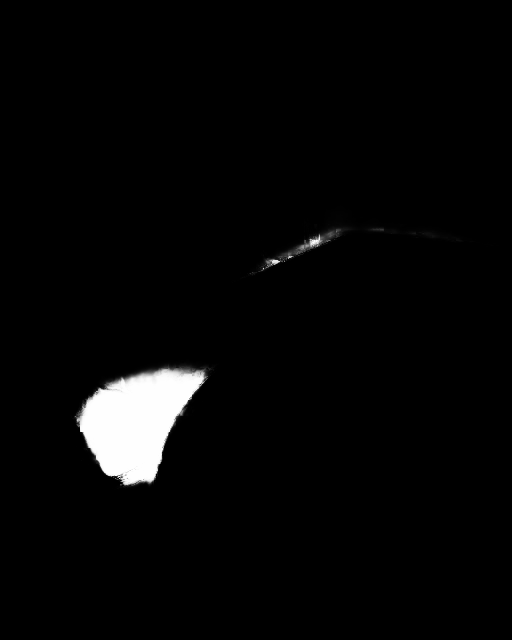}    
    \\
    \includegraphics[width=0.06\textwidth]{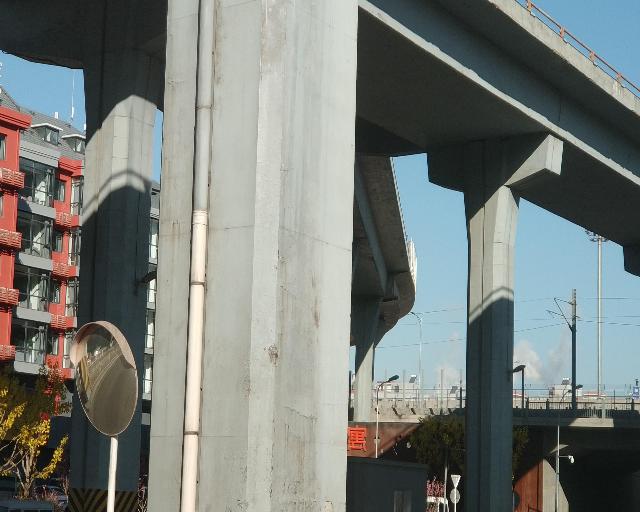}
    \includegraphics[width=0.06\textwidth]{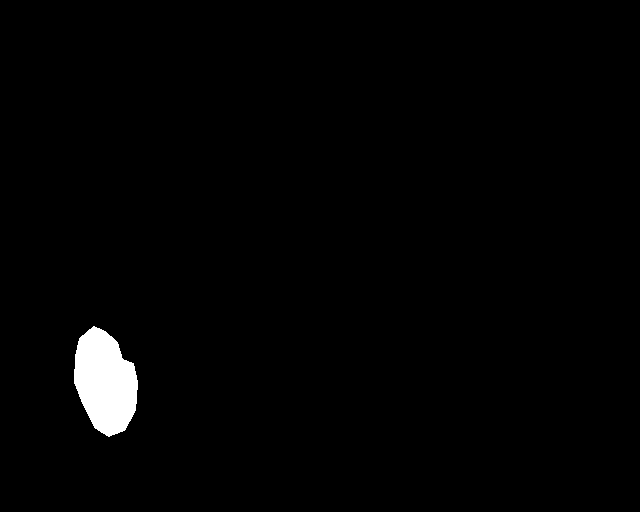}
    \includegraphics[width=0.06\textwidth]{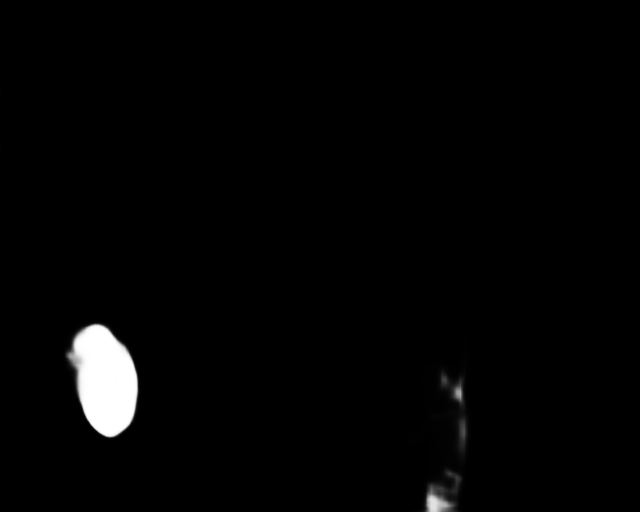}
    \includegraphics[width=0.06\textwidth]{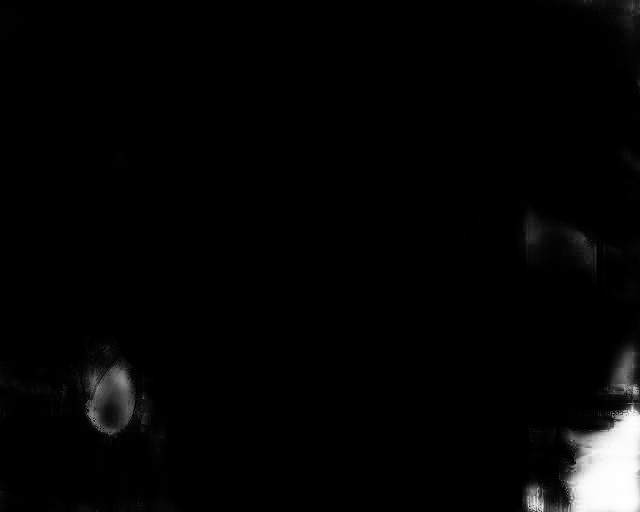}
    \includegraphics[width=0.06\textwidth]{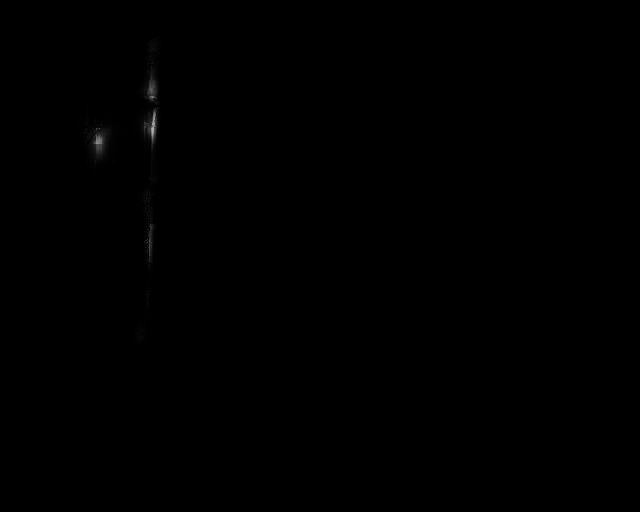}
    \includegraphics[width=0.06\textwidth]{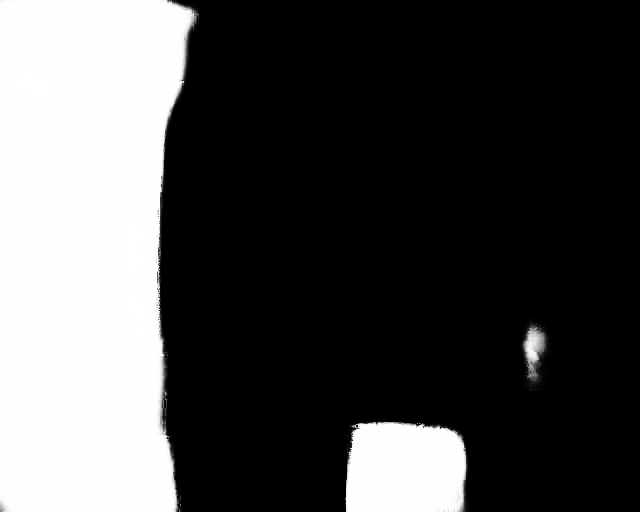}
    \includegraphics[width=0.06\textwidth]{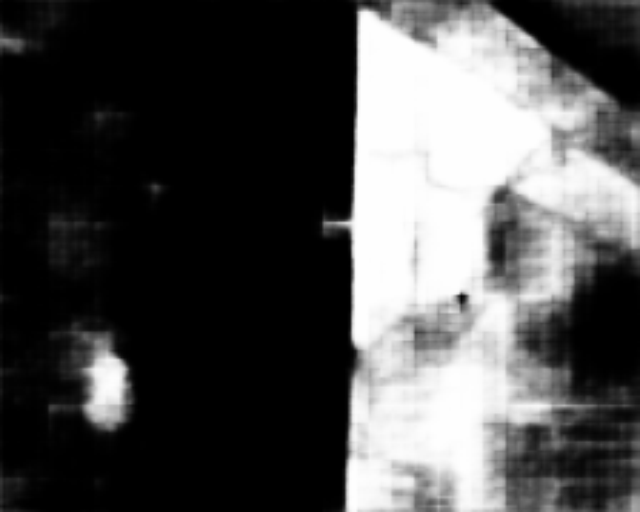}
    \includegraphics[width=0.06\textwidth]{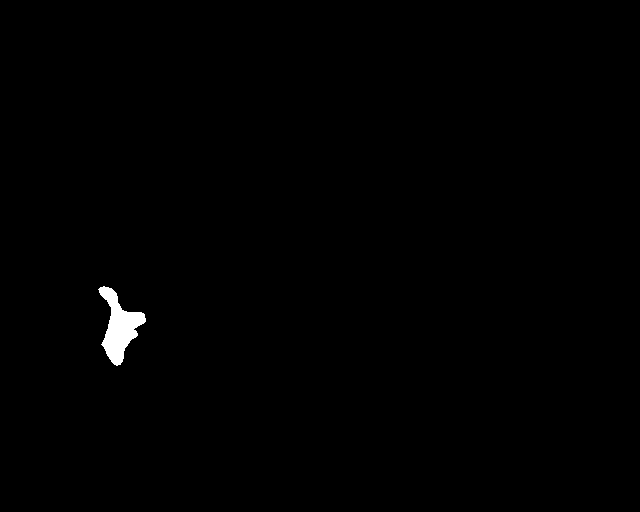}
    \includegraphics[width=0.06\textwidth]{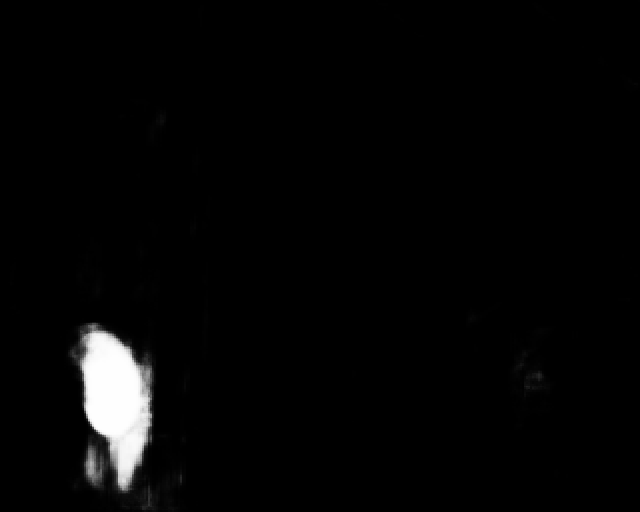}
    \includegraphics[width=0.06\textwidth]{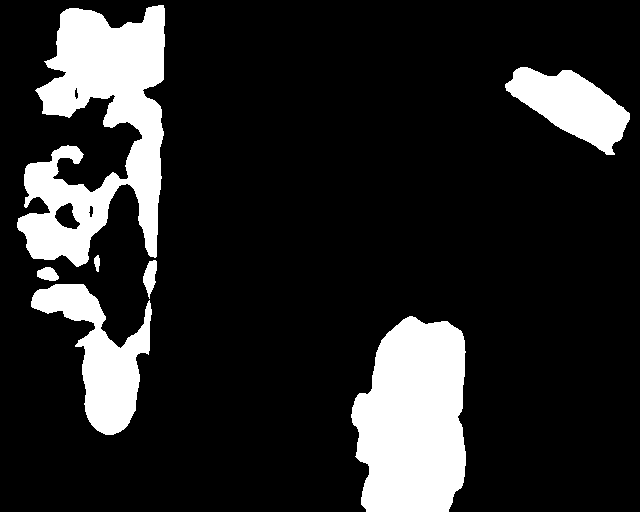}
    \includegraphics[width=0.06\textwidth]{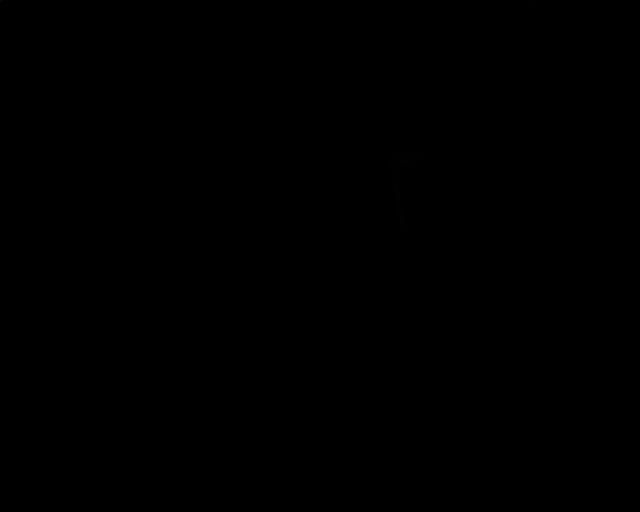}
    \includegraphics[width=0.06\textwidth]{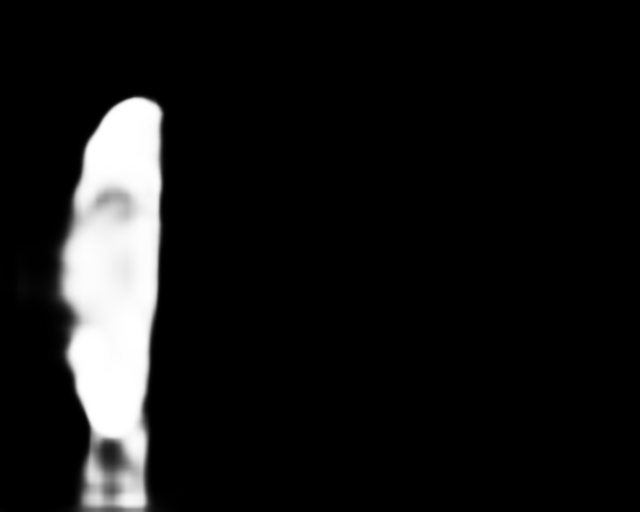}
    \includegraphics[width=0.06\textwidth]{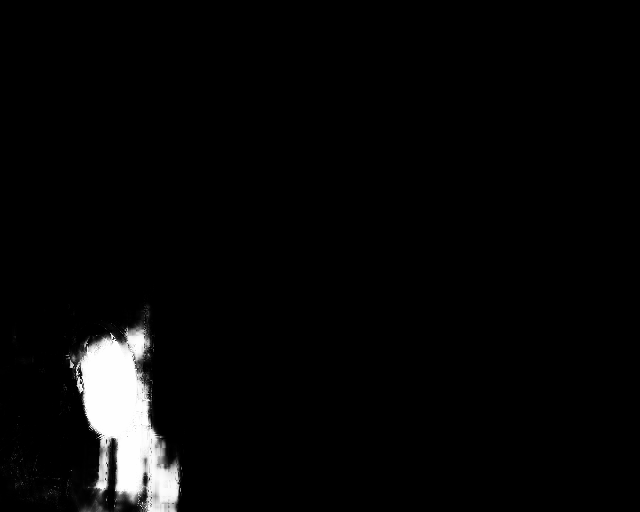}   
    \\
    \includegraphics[width=0.06\textwidth]{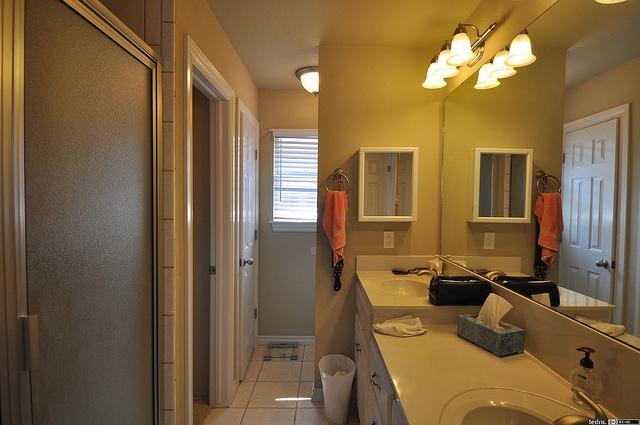}
    \includegraphics[width=0.06\textwidth]{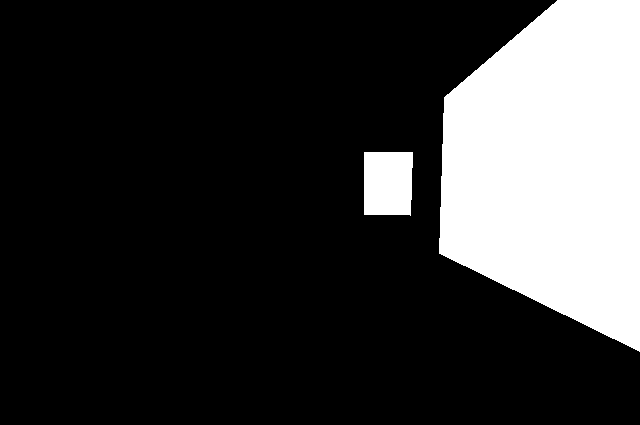}
    \includegraphics[width=0.06\textwidth]{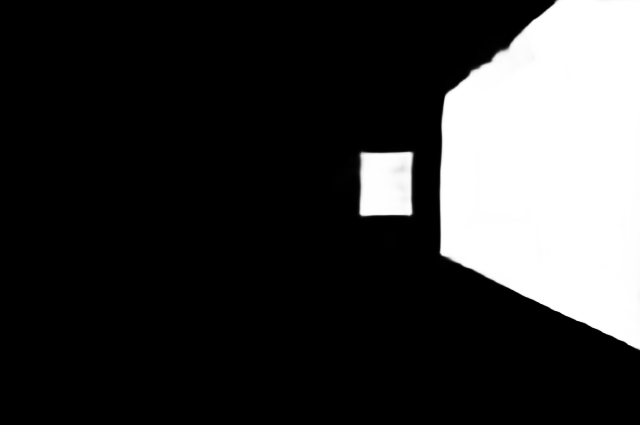}
    \includegraphics[width=0.06\textwidth]{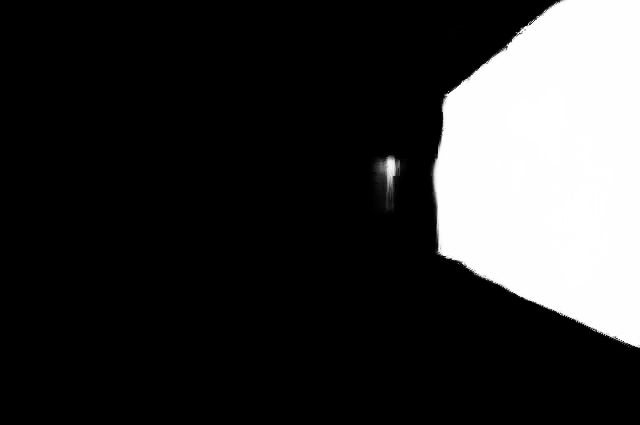}
    \includegraphics[width=0.06\textwidth]{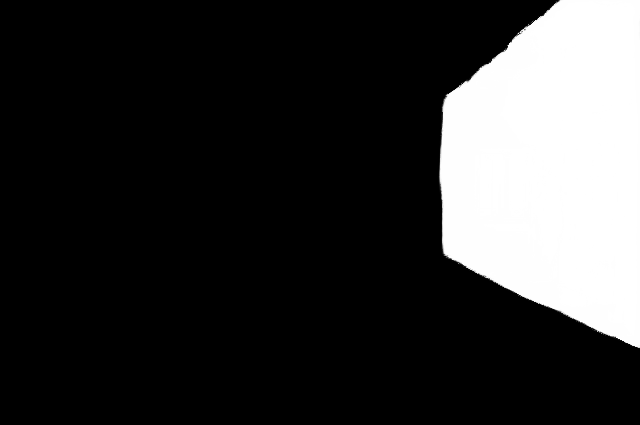}
    \includegraphics[width=0.06\textwidth]{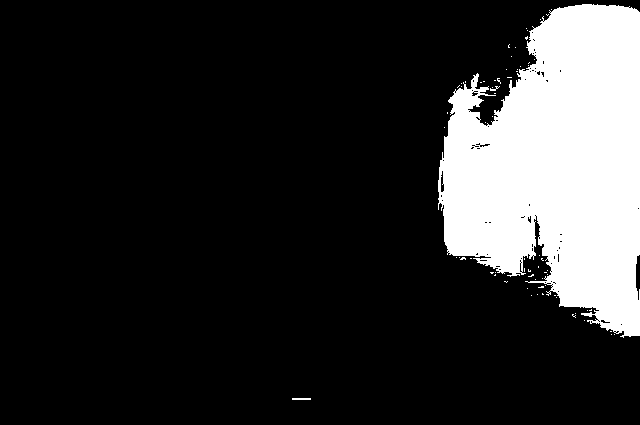}
    \includegraphics[width=0.06\textwidth]{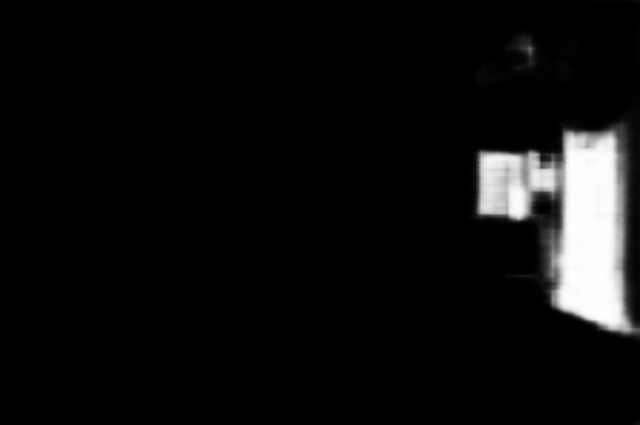}
    \includegraphics[width=0.06\textwidth]{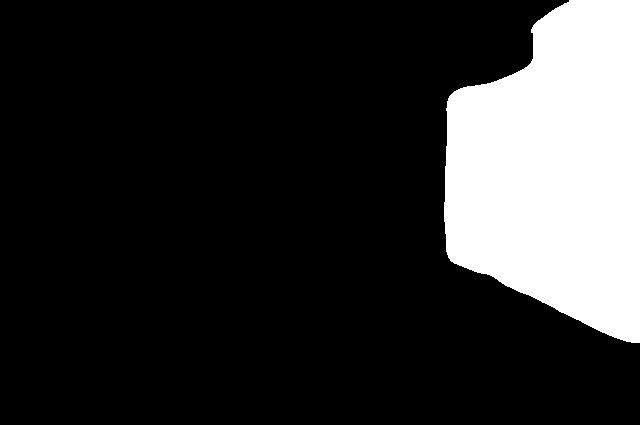}
    \includegraphics[width=0.06\textwidth]{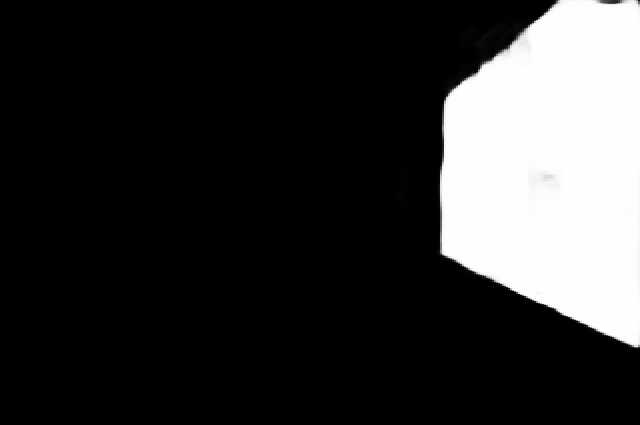}
    \includegraphics[width=0.06\textwidth]{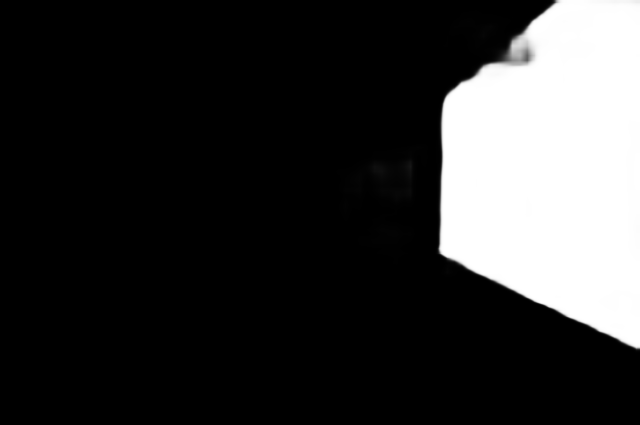}
    \includegraphics[width=0.06\textwidth]{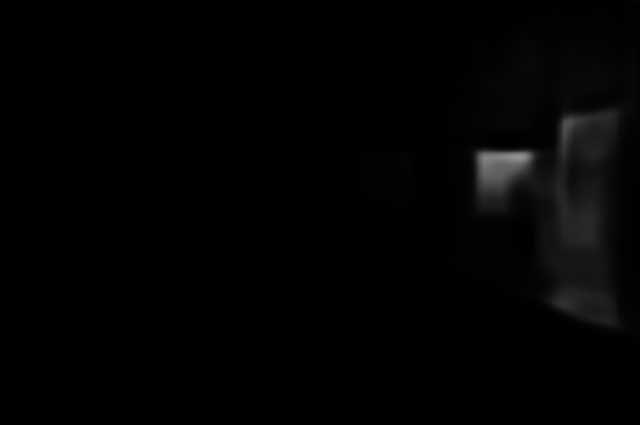}
    \includegraphics[width=0.06\textwidth]{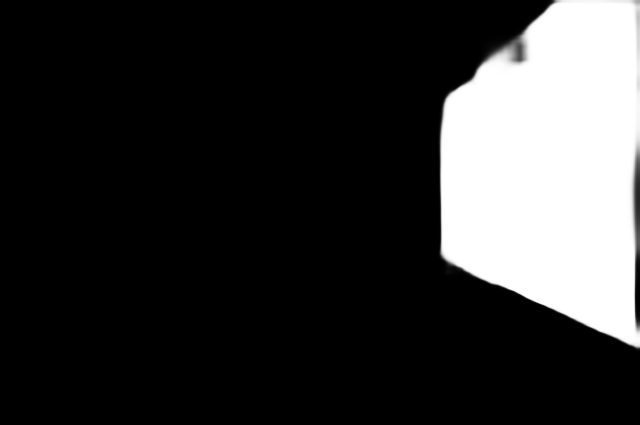}
    \includegraphics[width=0.06\textwidth]{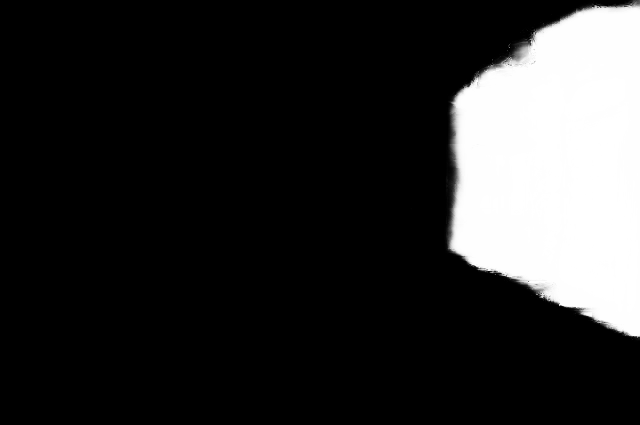}
    \\
    \includegraphics[width=0.06\textwidth]{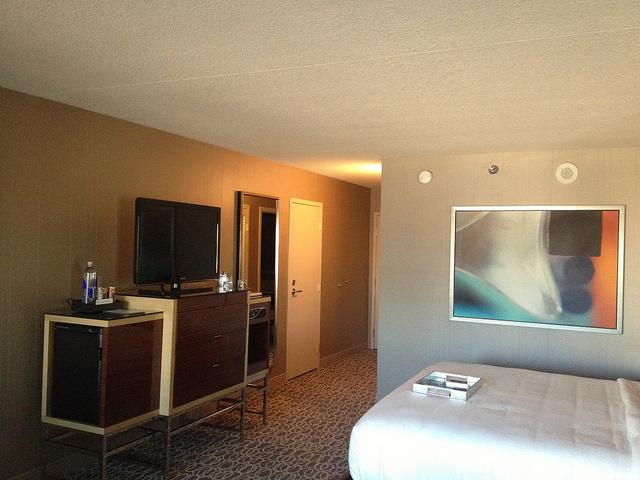}
    \includegraphics[width=0.06\textwidth]{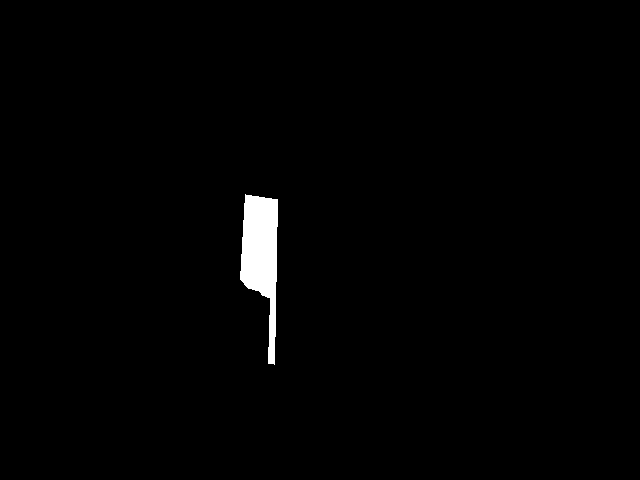}
    \includegraphics[width=0.06\textwidth]{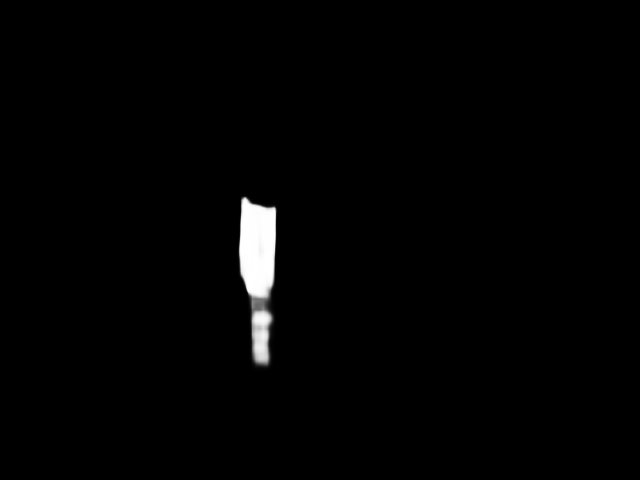}
    \includegraphics[width=0.06\textwidth]{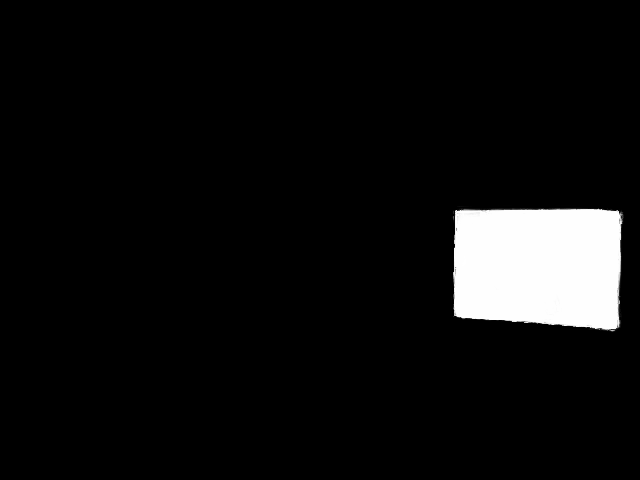}
    \includegraphics[width=0.06\textwidth]{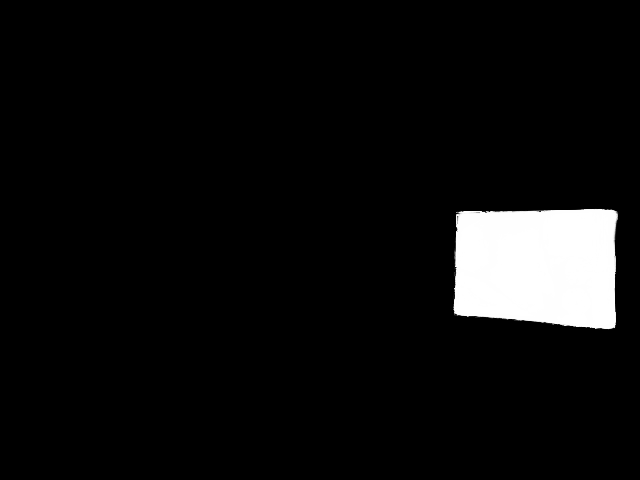}
    \includegraphics[width=0.06\textwidth]{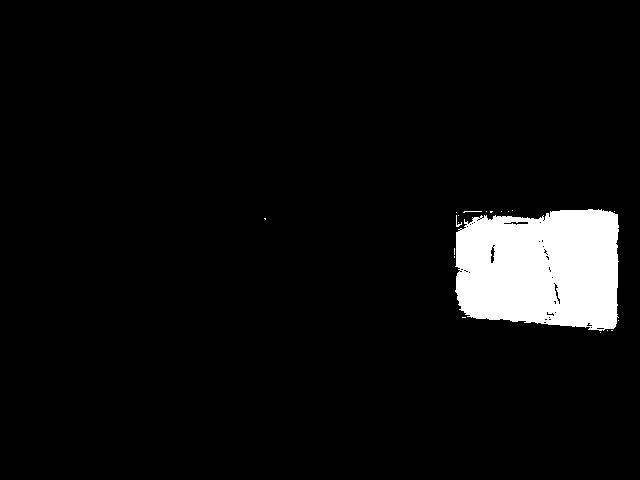}
    \includegraphics[width=0.06\textwidth]{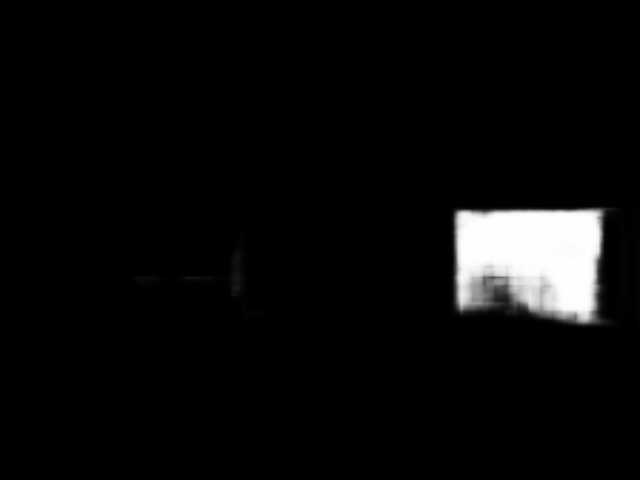}
    \includegraphics[width=0.06\textwidth]{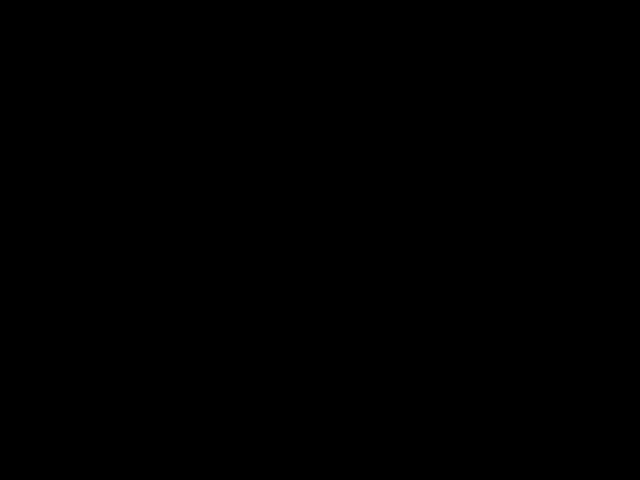}
    \includegraphics[width=0.06\textwidth]{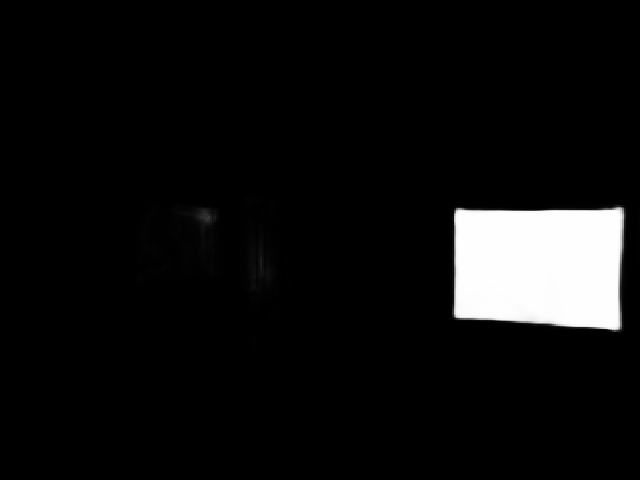}
    \includegraphics[width=0.06\textwidth]{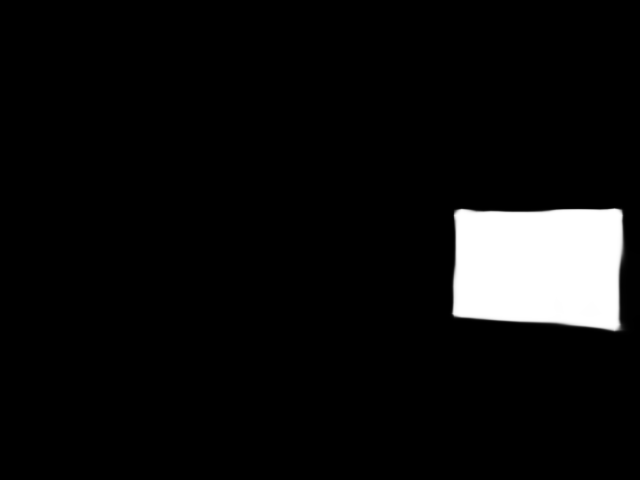}
    \includegraphics[width=0.06\textwidth]{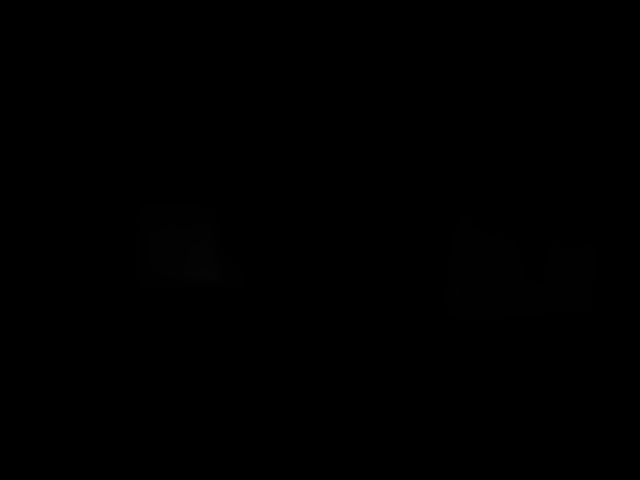}
    \includegraphics[width=0.06\textwidth]{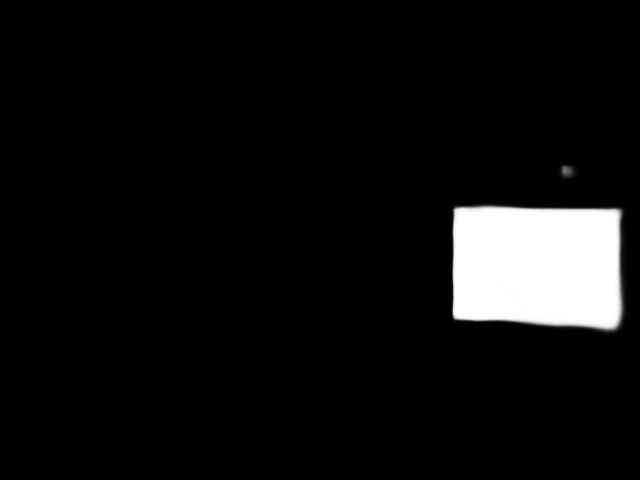}
    \includegraphics[width=0.06\textwidth]{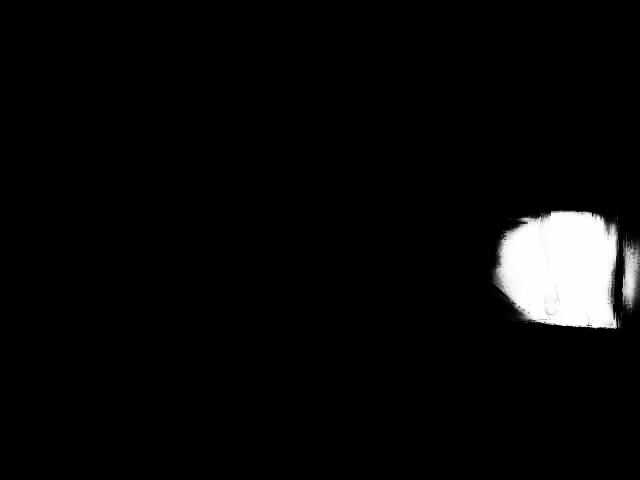}    
    \\
    \includegraphics[width=0.06\textwidth]{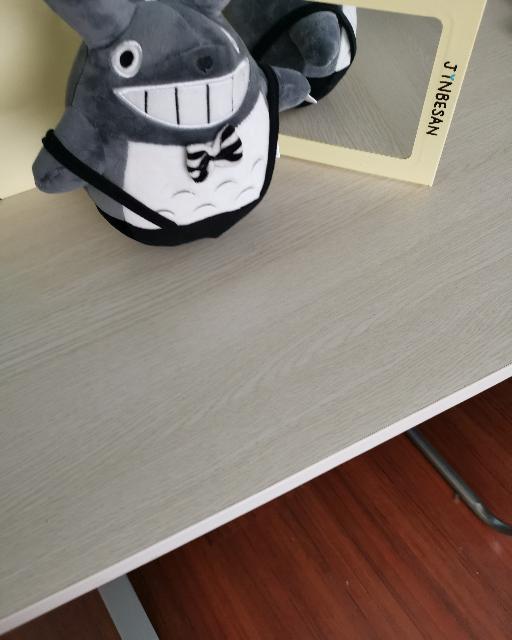}
    \includegraphics[width=0.06\textwidth]{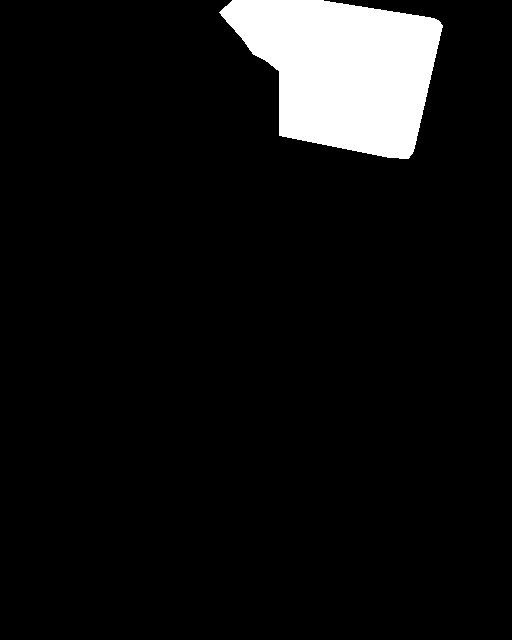}
    \includegraphics[width=0.06\textwidth]{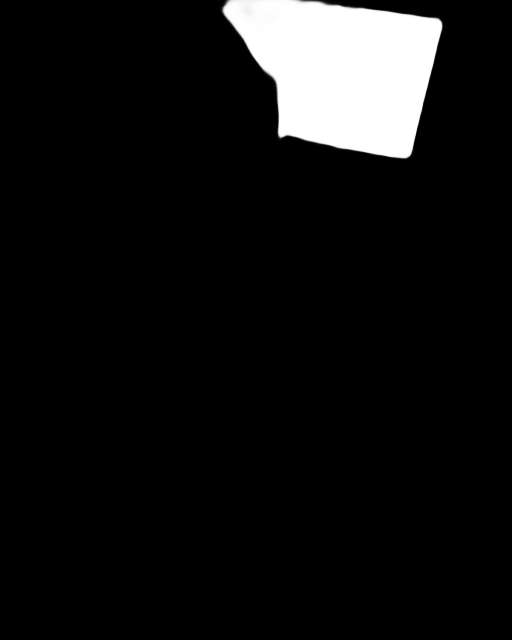}
    \includegraphics[width=0.06\textwidth]{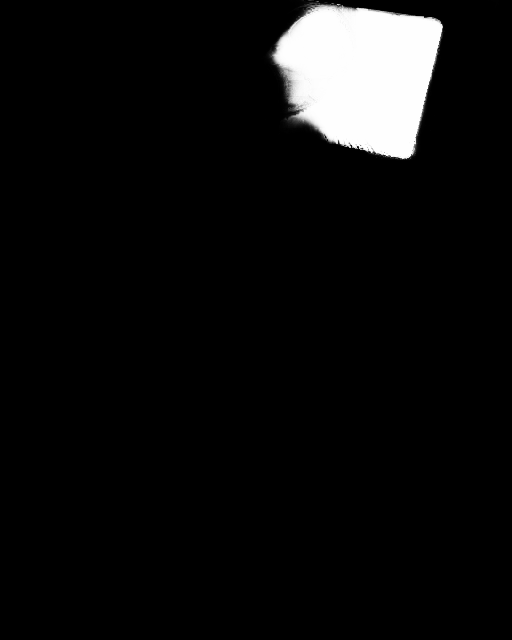}
    \includegraphics[width=0.06\textwidth]{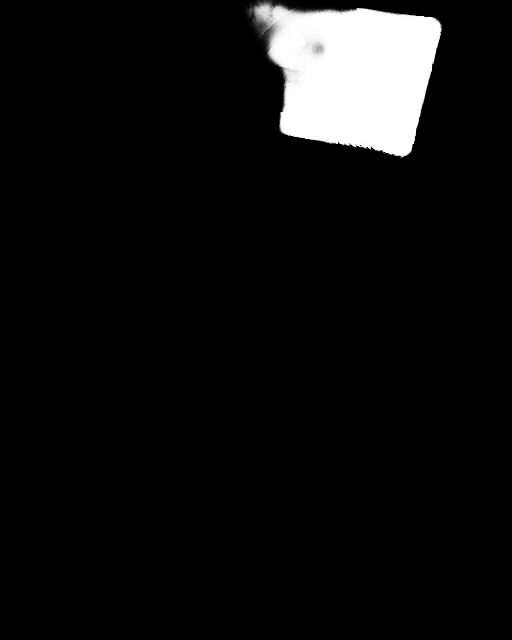}
    \includegraphics[width=0.06\textwidth]{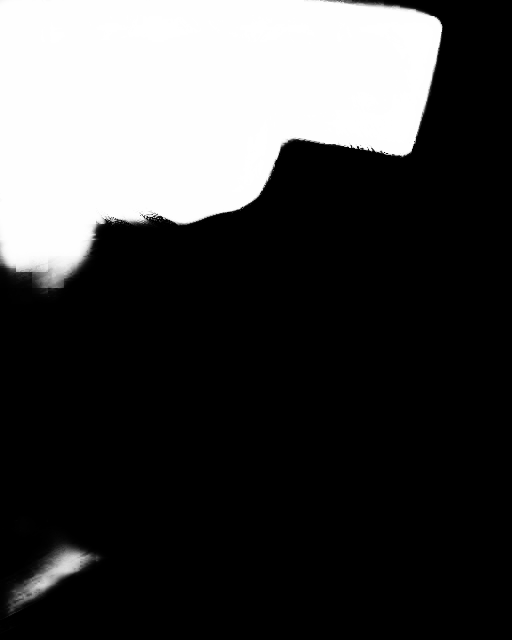}
    \includegraphics[width=0.06\textwidth]{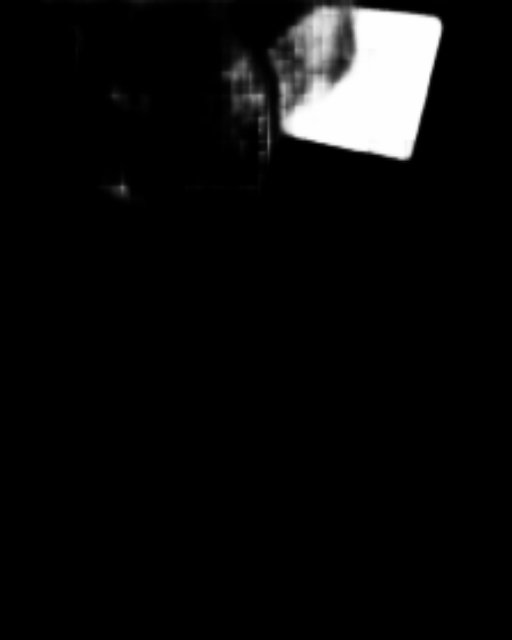}
    \includegraphics[width=0.06\textwidth]{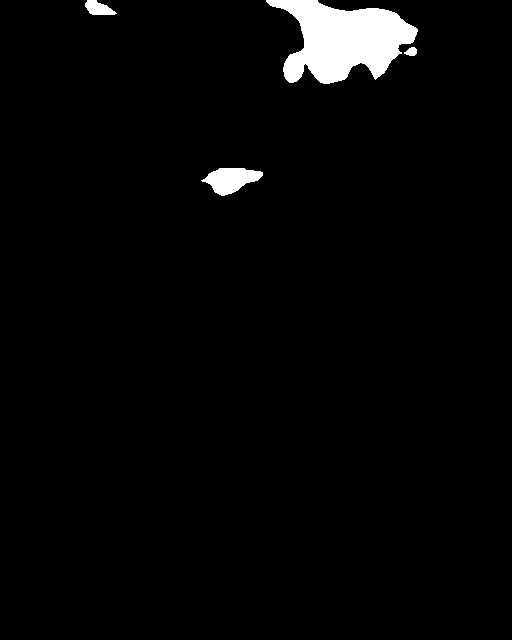}
    \includegraphics[width=0.06\textwidth]{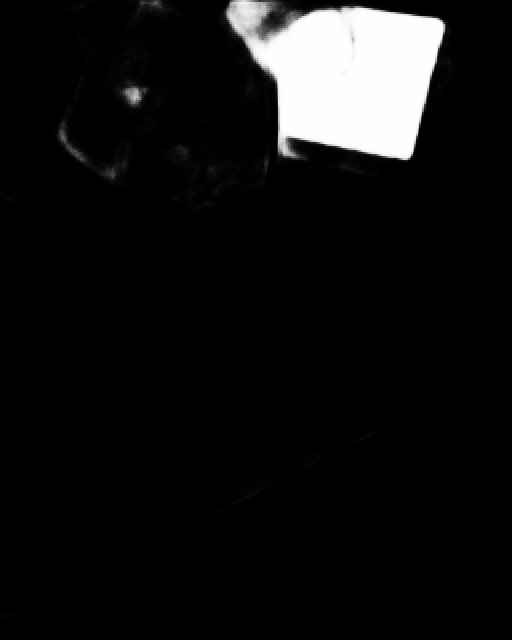}
    \includegraphics[width=0.06\textwidth]{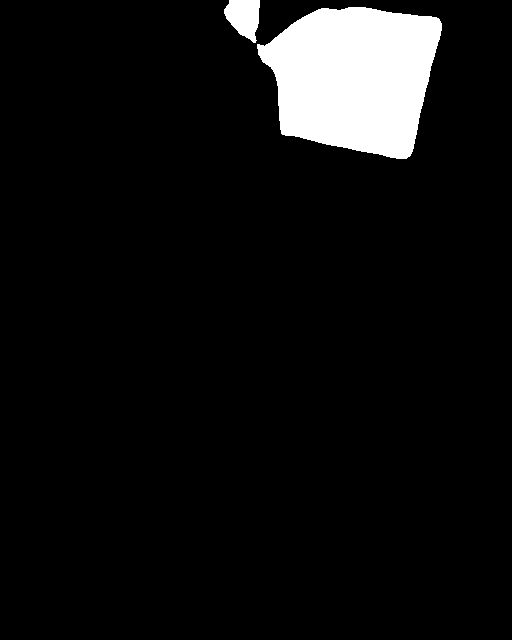}
    \includegraphics[width=0.06\textwidth]{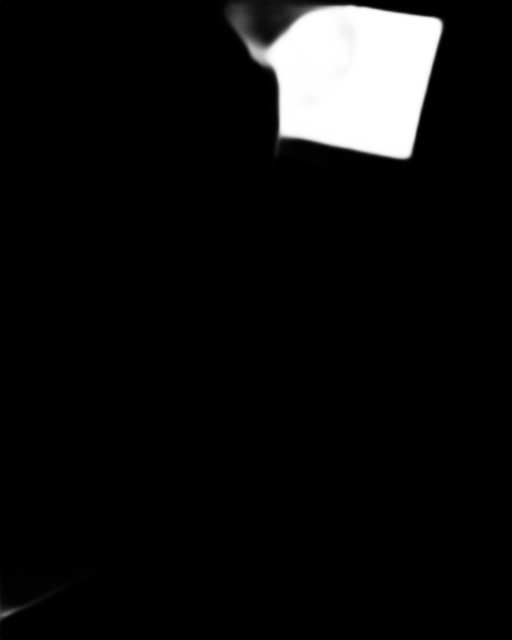}
    \includegraphics[width=0.06\textwidth]{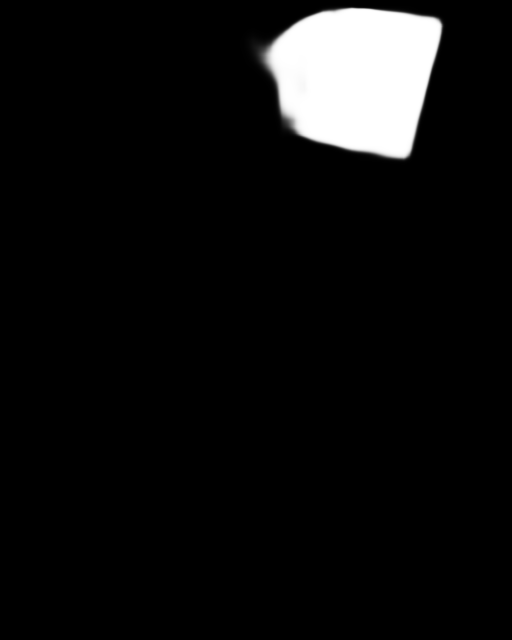}
    \includegraphics[width=0.06\textwidth]{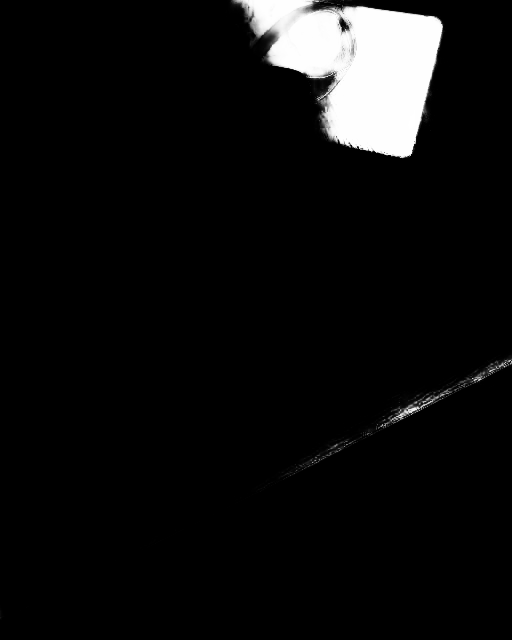}   
    \\
    
    \makebox[0.06\textwidth]{\scriptsize Input}
    \makebox[0.06\textwidth]{\scriptsize GT}
    \makebox[0.06\textwidth]{\scriptsize HetNet}
    \makebox[0.06\textwidth]{\scriptsize SANet}
    \makebox[0.06\textwidth]{\scriptsize PMD}
    \makebox[0.06\textwidth]{\scriptsize MirrorNet}
    \makebox[0.06\textwidth]{\scriptsize VST}
    \makebox[0.06\textwidth]{\scriptsize SETR}
    \makebox[0.06\textwidth]{\scriptsize MINet}
    \makebox[0.06\textwidth]{\scriptsize LDF}
    \makebox[0.06\textwidth]{\scriptsize EGNet}
    \makebox[0.06\textwidth]{\scriptsize CPDNet}
    \makebox[0.06\textwidth]{\scriptsize R$^{3}$Net}  
    \\
    \caption{Qualitative comparison of our model with relevant state-of-the-arts in challenging scenarios.} 
    \label{fig:maps}
\end{figure*}

We provide some visual comparisons with state-of-the-art methods. As shown in Figure~\ref{fig:maps}, our method can generate more precise segmentation than other counterparts. It performs excellently in various challenging scenarios, such as high-intrinsic similar surroundings (rows 1, 2), mirrors split by objects (row 3), ambiguous regions outside mirrors (row 4), tiny mirrors (rows 5, 6), and partially hidden mirrors (rows 7, 8). Note that we do not use any post-processing to generate our maps. This shows that our method is effective and robust in processing complex images.

\subsection{Ablation Study}
To better analyze the architecture and effectiveness of our network, we conduct ablation studies of each component in our proposed network on the MSD dataset.

\subsubsection{Network Architecture Analysis.}
In this section, we focus on analyzing the network structure to demonstrate the rationality and necessity of learning specific information with the heterogeneous modules at different stages. As shown in Table~\ref{tab:structure}, using MICs at all stages (A$_{a}$) performs the worst. Models with RSLs at shallow stages and MICs at deep stages (A$_{ba}$) or RSLs (A$_{b}$) at all stages have similar performances, which are not as effective as our HetNet.
Low-level features focus more on colors, shapes, texture and contain more precise spatial information, while high-level features involve more semantics with rough spatial information.
MIC aims to learn low-level contrasts, but RSL extracts high-level understandings. Hence, it is more reasonable to learn low-level information to roughly localize mirrors with MICs at shallow (1-3) stages and learn high-level information by RSLs at deep (4-5) stages.

\begin{table}[t]
\centering
\caption{The ablation study results of network architecture. A$_{ba}$, A$_{a}$, A$_{b}$ denote applying RSL at the shallow stages and MIC at the deep stages, MIC at all stages, and RSL at all stages, respectively.}
\scalebox{0.7}{
\begin{tabular}{c|ccc|ccc}
\hline
Architecture & MAE$\downarrow$   & IoU$\uparrow$   & F$_{\beta}\uparrow$ & Para. & FLOPs & FPS   \\ \hline
A$_{ba}$  & 0.046 & 0.821 & 0.897 & 50.13 & 60.89 & 37.73\\
A$_{a}$   & 0.049 & 0.811 & 0.889 & 47.58 & 26.90 & 43.34 \\
A$_{b}$   & 0.046 & 0.817 & 0.897 & 52.50 & 61.67 & 39.00 \\ \hline
HetNet   & 0.043 & 0.828 & 0.906 & 49.92 &27.69 &49.23\\ \hline
\end{tabular}
}

\label{tab:structure}
\end{table}

\subsubsection{Component Analysis.}
To verify the component effectiveness, we conduct ablation experiments by gradually adding them \ryn{to the network}. We simply run the original backbone network \cite{xie2017aggregated} (I) with the remaining HetNet architecture without the 6th stage as a baseline, and then insert GE (II) into it to complete 6 stages. We gradually add three alternative components to it. The first one adds one ICFE (III) instead of MIC. The second includes the entire MIC component (IV). Both of the above two methods exclude RSLs. The third one applies RSLs without MICs (V).

\begin{figure}[t] \centering
    \includegraphics[width=0.055\textwidth]{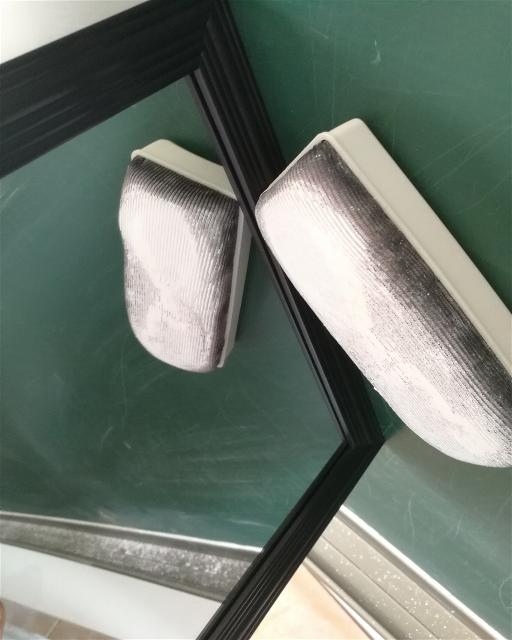}
    \includegraphics[width=0.055\textwidth]{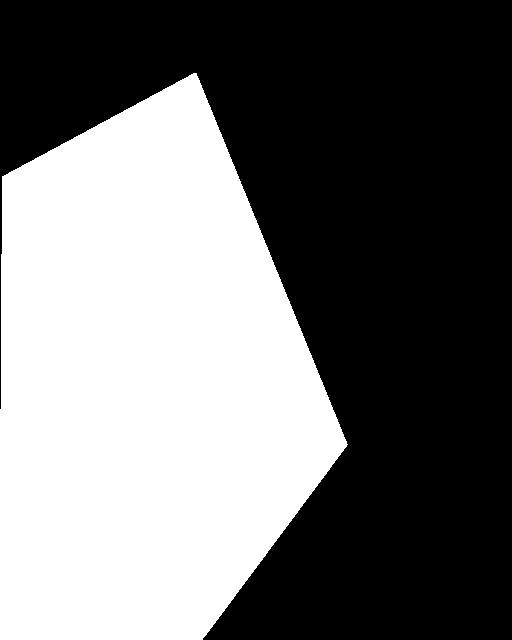}
    \includegraphics[width=0.055\textwidth]{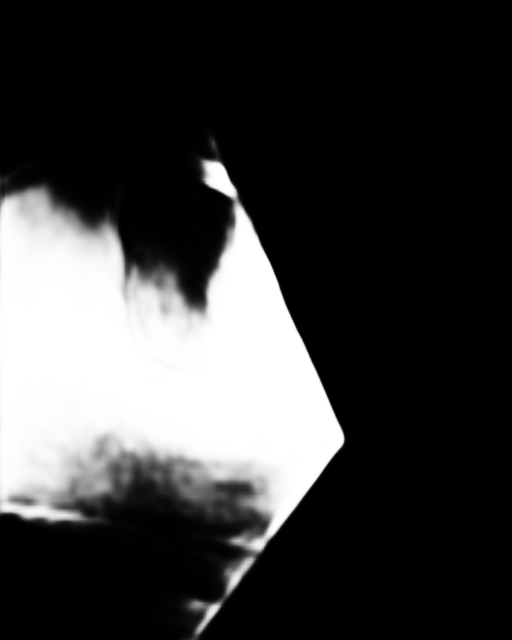}
    \includegraphics[width=0.055\textwidth]{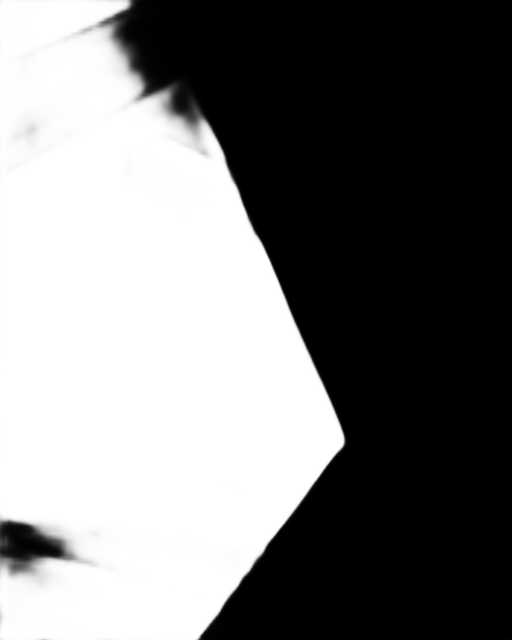}
    \includegraphics[width=0.055\textwidth]{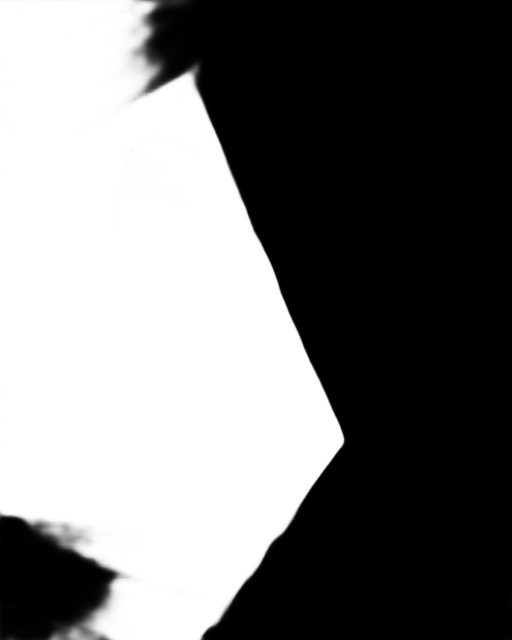}
    \includegraphics[width=0.055\textwidth]{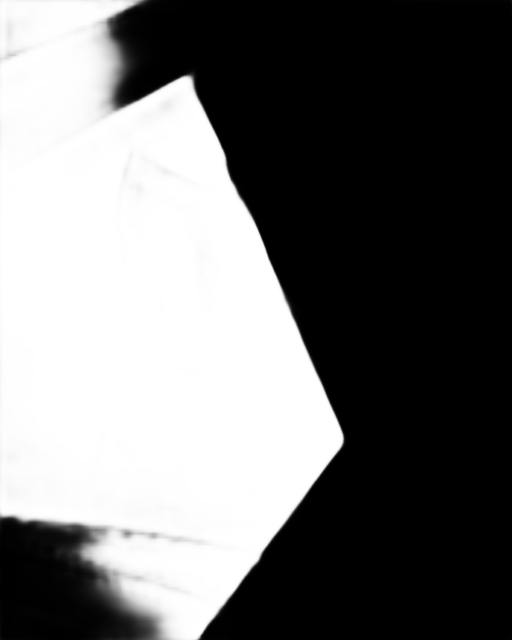}
    \includegraphics[width=0.055\textwidth]{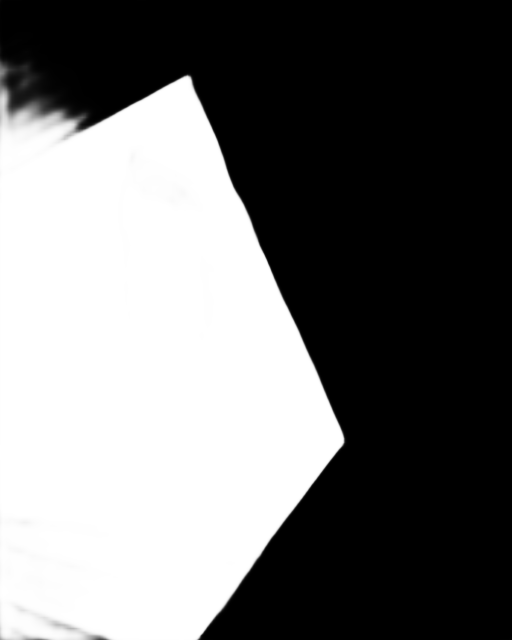}
    \includegraphics[width=0.055\textwidth]{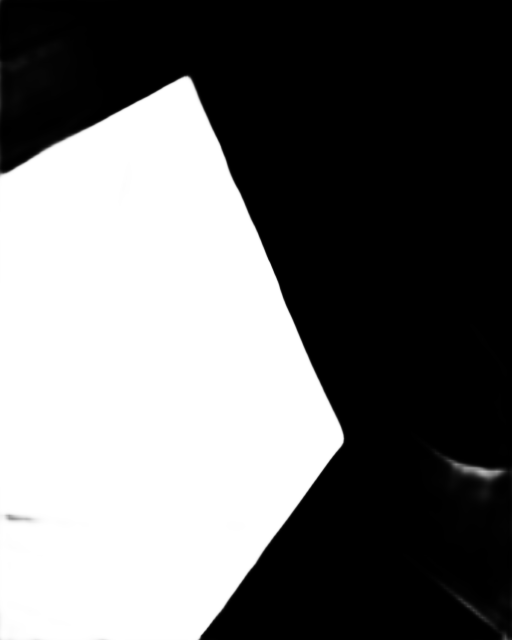}
    \\
    \makebox[0.055\textwidth]{\scriptsize Input}
    \makebox[0.055\textwidth]{\scriptsize GT}
    \makebox[0.055\textwidth]{\scriptsize (I)}
    \makebox[0.055\textwidth]{\scriptsize (II)}
    \makebox[0.055\textwidth]{\scriptsize (III)}
    \makebox[0.055\textwidth]{\scriptsize (IV)}
    \makebox[0.055\textwidth]{\scriptsize (V)}
    \makebox[0.055\textwidth]{\scriptsize HetNet}
    \\
    \caption{A visual example of the ablation study. (I) to (V) correspond to the prediction maps from five ablated models: “basic”, “basic + GE”, “basic + GE + ICFEs”, “basic + GE + MICs”, “basic + GE + RSLs”, respectively.} 
    \label{fig:ablation}
\end{figure}

Table~\ref{tab:component} illustrates the experimental results. The ablated model (I) performs the worst of all. We may also observe that adding MICs (IV) or RSLs (V) is generally better than the other alternative (\textit{i.e.}, III). As MICs learn low-level information in two orientations instead of a single orientation, while (IV) performs better than (III). In addition, “basic + GE + MICs” (IV) shows a slight overall advantage over “basic + GE + RSLs” (V), but when MICs and RSLs are combined (HetNet or “basic + GE + MICs + RSLs”), it outperforms all the other ablated models. It proves that the cooperation of low- and high-level cues is more effective than single-level understandings in mirror detection.
Figure~\ref{fig:ablation} shows a visual example where MICs and RSLs play an important role together.

\begin{table}[t]
\centering
\caption{The ablation study results of components. By adding each component gradually, our model achieves the best performance.}
\scalebox{0.6}{
\begin{tabular}{c|ccccc|ccc|ccc}
\hline
Ablation & Base       & GE         & ICFEs       & MICs        & RSLs        & MAE$\downarrow$   & IoU$\uparrow$   & F$_{\beta}\uparrow$ &  Para. & FLOPs & FPS    \\ \hline
I &$\surd$ &            &            &            &            & 0.056 & 0.773 & 0.872 &47.53 &26.61 &60.31\\
II &$\surd$ & $\surd$ &            &            &            & 0.050 & 0.805 & 0.882 &47.53 &26.68 &58.05 \\
III &$\surd$ & $\surd$ & $\surd$ &            &            & 0.053 & 0.781 & 0.878 &47.55 &26.68 &53.87\\
IV &$\surd$ & $\surd$ &            & $\surd$ &            & 0.049 & 0.811 & 0.892 &47.56 &26.68 &51.45\\
V &$\surd$ & $\surd$ &            &            & $\surd$ & 0.049 & 0.809 & 0.890 &49.92 &27.47 &54.20\\ \hline
HetNet &$\surd$ & $\surd$ &            & $\surd$ & $\surd$ & 0.043 & 0.828 & 0.906 &49.92 &27.69 &49.23\\ \hline
\end{tabular}
}
\label{tab:component}
\end{table}

\subsubsection{Effectiveness of the Rotation Strategy in the MIC Module.}
In Table~\ref{tab:MIC}, we compare five rotation strategies in MIC and show the effectiveness of our multi-orientation strategy. 
MIC performing better than 1 ICFE (“ICFE”) or 2 parallel ICFEs (“ICFE+ICFE”). This shows that rotation helps learn more comprehensive low-level information in two orientations.
To obtain as much distinct information as possible, we adopt an equal division strategy for orientations. If we use lines to denote orientations on a 2D plane, we expect intersecting lines to divide 360$^{\circ}$ equally, \textit{e.g.}, 1 line into 2*180$^{\circ}$, 2 lines into 4*90$^{\circ}$, and 3 lines into 6*60$^{\circ}$. However, rotating tensors for non-(90*k)$^{\circ}$ induces information loss. For example, if a tensor rotates 60$^{\circ}$, we need to add paddings or crop it. Hence, 90*k$^{\circ}$ are better options.
The failure of ICFE*3 is possibly caused by observing one orientation (0$^{\circ}$,180$^{\circ}$) twice, which makes information on this orientation stronger than the other (90$^{\circ}$). ICFE*4 performs better than ICFE*3 for its balanced observation on two orientations. However, it may introduce more noise so that it is worse than our HetNet (MIC). 

\begin{table}[h]
\centering
\caption{The ablation study results of MIC rotation strategies. ICFE, ICFE+ICFE, ICFE*3, ICFE*4, and MIC denote a single ICFE, 2 same-orientation ICFEs, 3 ICFEs with ($0^{\circ},90^{\circ},180^{\circ}$) orientations, 4 ICFEs with ($0^{\circ},90^{\circ}, 180^{\circ},270^{\circ}$), and 2 ICFEs with ($0^{\circ},180^{\circ}$) orientations, respectively.}
\scalebox{0.7}{
\begin{tabular}{c|ccc|ccc}
\hline
Strategy  & MAE$\downarrow$   & IoU$\uparrow$   & F$_{\beta}\uparrow$ & Para. & FLOPs & FPS    \\ \hline
ICFE      & 0.049 & 0.802 & 0.887 & 49.92 & 27.68 & 49.52 \\
ICFE+ICFE & 0.048 & 0.806 & 0.891 &49.92 &27.69 &45.58 \\ 
ICFE*3    & 0.047 & 0.821 & 0.895 &49.93 &27.69 &40.73 \\
ICFE*4    & 0.046 & 0.820 & 0.901 &49.93 &27.70 &36.08 \\ \hline
MIC       & 0.043 & 0.828 & 0.906 &49.92 &27.69 &49.23 \\ \hline
\end{tabular}
}

\label{tab:MIC}
\end{table}

\section{Conclusions}
In this paper, we propose a highly efficient mirror detection model HetNet. To meet the trade-off of efficiency and accuracy, we adopt heterogeneous modules at different stages to provide the benefits of feature characteristics at both low and high levels. Additionally, considering the low-level contrasts inside and outside mirrors, we propose a multi-orientation intensity-based contrasted (MIC) module to learn low-level understanding in two orientations to select likely mirror regions. To further confirm mirrors, we propose a reflection semantic logical (RSL) module to extract high-level information. 
Overall, HetNet has outstanding and efficient feature extraction performances, making it effective and robust in challenging scenarios. 
Experimental results on two benchmarks illustrate that our method outperforms SOTA methods on three evaluation metrics with an average enhancement of 8.9$\%$ on MAE, 3.1$\%$ on IoU, 2.0$\%$ on F-measure, as well as 72.73$\%$ fewer model FLOPs and 664$\%$ faster than the SOTA mirror detection method PMD~\cite{lin2020progressive}.

\begin{figure}[h] \centering
    \includegraphics[width=0.15\textwidth]{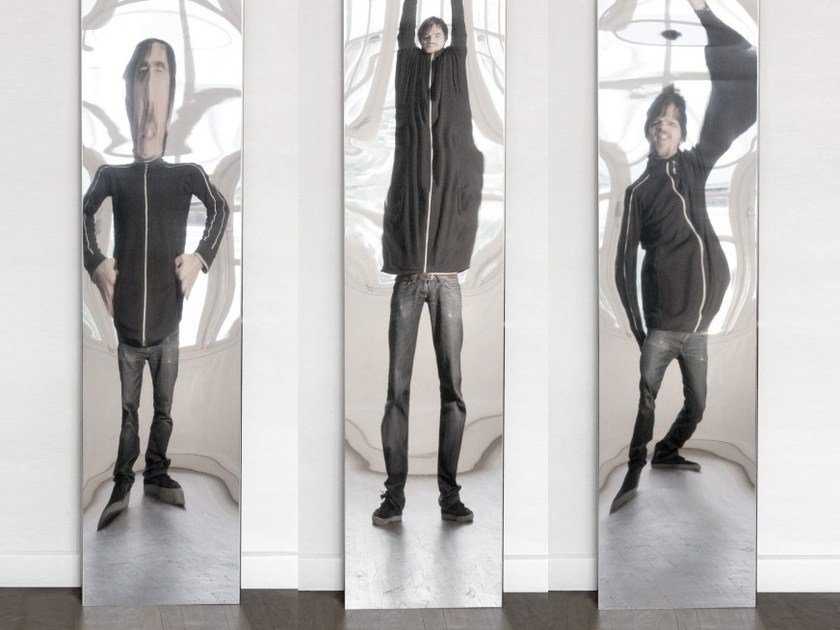}
    \includegraphics[width=0.15\textwidth]{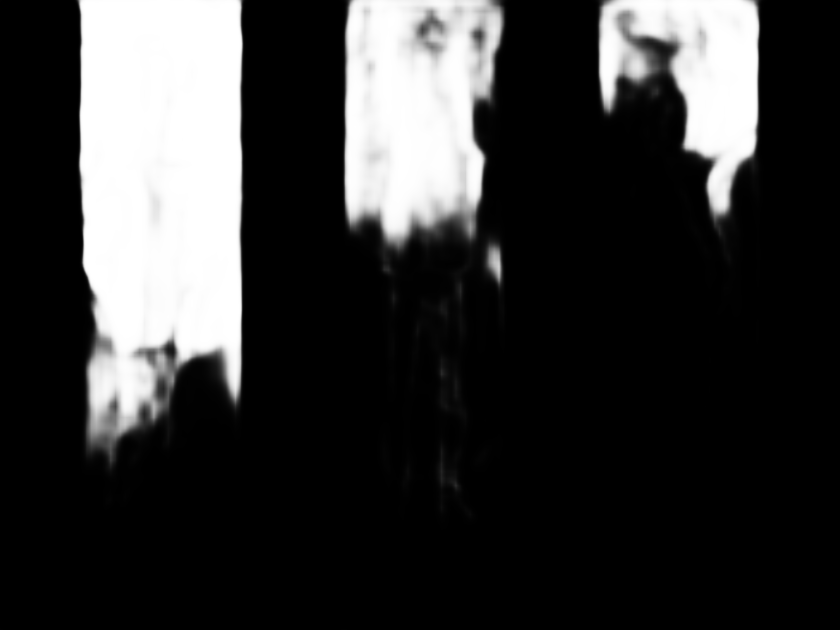}
    \caption{Failure cases. Our model may fail in scenarios with complex reflection mirrors, like distorting mirrors.} 
    \label{fig:limitation}
\end{figure}

Our method does have limitations. Since both MSD and PMD datasets collect mostly regular mirrors, our method may fail in some mirrors with special reflection property occasions. In Figure~\ref{fig:limitation}, the three mirrors reflect the same man with three different statuses. In the right image, the mirrors show complex intensity contrast. Hence, the distortion of high-level or low-level information is even more challenging. For future work, we are currently considering additional information to help detect different kinds of mirrors.

\bibliography{aaai23}

\end{document}